\documentclass[journal]{IEEEtran}

\usepackage{amsmath}
\usepackage{booktabs}
\usepackage{xfrac}
\usepackage{graphicx}
\usepackage{tikz}
\usepackage{gensymb}
\usepackage{caption}
\usepackage{subcaption}
\usepackage{multirow}
\usepackage{ulem}
\usepackage{newtxmath}
\usepackage{xcolor,colortbl}
\usepackage{authblk}
\usepackage{textcomp}
\begin{document}

\title{Design and integration of a parallel, soft robotic end-effector for extracorporeal ultrasound}

\author{Lukas~Lindenroth, 
        Richard James Housden, Shuangyi Wang, Junghwan Back, Kawal Rhode
        and~Hongbin~Liu 

\thanks{L. Lindenroth is with the Department of Engineering, King's College London, UK.}
\thanks{R. J. Housden, S. Wang, J. Back, K. Rhode and H. Liu are with the School of Biomedical Engineering and Imaging Sciences, King’s College London, UK.}
\thanks{Corresponding author: Hongbin Liu (hongbin.liu@kcl.ac.uk)}
\thanks{\textcopyright 2019 IEEE. Personal use of this material is permitted. Permission from IEEE must be obtained for all other uses, in any current or future media, including reprinting/republishing this material for advertising or promotional purposes, creating new collective works, for resale or redistribution to servers or lists, or reuse of any copyrighted component of this work in other works.}
}

\maketitle

\begin{abstract}
 
Objective: In this work we address limitations in state-of-the-art ultrasound robots by designing and integrating a novel soft robotic system for ultrasound imaging. It employs the inherent qualities of soft fluidic actuators to establish safe, adaptable interaction between ultrasound probe and patient. Methods: We acquire clinical data to determine the movement ranges and force levels required in prenatal foetal ultrasound imaging and design the soft robotic end-effector accordingly. We verify its mechanical characteristics, derive and validate a kinetostatic model and demonstrate controllability and imaging capabilities on an ultrasound phantom. Results: The soft robot exhibits the desired stiffness characteristics and is able to reach 100\% of the required workspace when no external force is present, and 95\% of the workspace when considering its compliance. The model can accurately predict the end-effector pose with a mean error of $\mathbf{1.18\pm0.29}$mm in position and $\mathbf{0.92\pm0.47\degree}$ in orientation. The derived controller is, with an average position error of 0.39mm, able to track a target pose efficiently without and with externally applied loads. Ultrasound images acquired with the system are of equally good quality compared to a manual sonographer scan. Conclusion: The system is able to withstand loads commonly applied during foetal ultrasound scans and remains controllable with a motion range similar to manual scanning. Significance: The proposed soft robot presents a safe, cost-effective solution to offloading sonographers in day-to-day scanning routines. The design and modelling paradigms are greatly generalizable and particularly suitable for designing soft robots for physical interaction tasks.
\end{abstract}


\IEEEpeerreviewmaketitle

\section{Introduction}

\IEEEPARstart{I}{t} is commonly accepted that sonographers are exposed to an increased risk in repetitive strain injury \cite{Seto2008, Janga2012, Harrison2015}. A representative study amongst diagnostic medical sonographers and vascular technologists indicates that a significant majority of sonographers experience pain while performing ultrasound scans \cite{Evans2009}. This suggests a high demand to improve ergonomics and offload sonographers during clinical scan procedures. 
Recent investigations show that besides diagnostic sonography, there is an increased demand for intraoperative transthoracic \cite{Ben-Dor2006, Hori2015} and transoesophegal \cite{Shanewise1999} ultrasound imaging, particularly for cardiac and lung procedures. Sonographers performing intraoperative ultrasound in for example cardiac catheterization procedures have therefore presumably an increased risk of radiation exposure \cite{McIlwain2014}.

\begin{figure}[t!]
\center{\includegraphics[width=\linewidth]{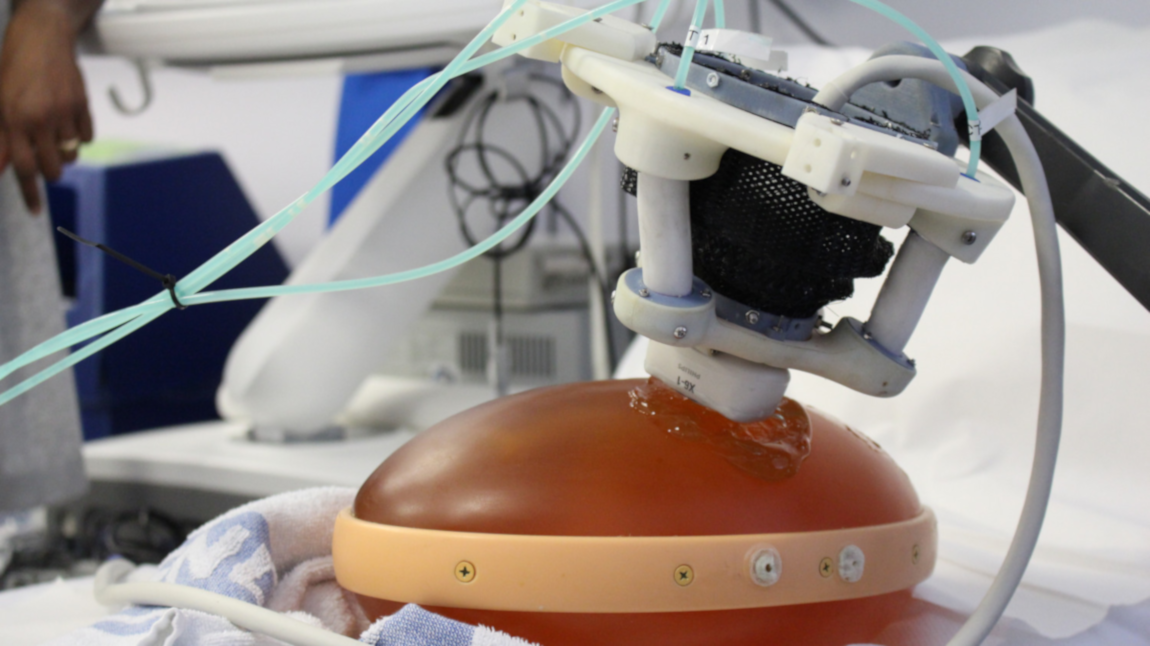}\caption{Soft robotic end-effector (SEE) performing ultrasound scan on abdominal prenatal phantom}}
\end{figure}
Automating diagnostic and intraoperative ultrasound procedures through robot-guidance or -assistance can help address the aforementioned problems and lay the groundwork for more intelligent image acquisition. Robotic ultrasound guidance has found particular application in procedures involving  steering orthopaedic \cite{Goncalves2014} or minimally-invasive surgical tools \cite{Antico2019UltrasoundProcedures} and biopsy needles \cite{Mahmoud2018EvolutionSystems}. Various robotic hardware solutions have been proposed. Researchers have adopted robotic platforms originally aimed at collaborative scenarios in industrial settings, such as Universal Robot’s UR-series \cite{Mathiassen2016, Sen2016} or the KUKA LWR \cite{Goncalves2014} and LBR iiwa \cite{Kojcev2016, Zettinig2016}. A commercial robotic manipulator has been released (LBR Med, KUKA AG, Augsburg, Germany) which is suitable for use in clinical environments due to its conformity with medical device safety (ISO 60601) and medical software regulations (ISO 62304). Current research suggests that such robots can be applied in diagnostics to autonomously perform aorta measurements \cite{Virga2016}, in combination with previously acquired MRI scans to autonomously find standard view-planes \cite{Hennersperger2017} and in intraoperative procedures to autonomously track surgical tools \cite{Salcudean2013}, amongst others. 
Whilst such robotic platforms allow for great flexibility through a large workspace and high manipulability, the use of large-scale robotic manipulators can pose various disadvantages for clinical integration. Diagnostic ultrasound scans are divided into their respective body area of interest. For an individual procedure such as a lower abdominal ultrasound scan, a robotic system is therefore only required to achieve a workspace to cover a fraction of the human body. This yields that common robotic manipulators could be oversized for such applications, which unnecessarily poses risks to patient safety. Despite high degrees of electrical safety, a mechanical system with a high mass can potentially be more dangerous \cite{Haddadin2008}.

To address this issue, researchers developed customized solutions which are tailored to the application-specific requirements of diagnostic and interventional sonography. Researchers \cite{Salcudean1999, Salcudean1999a, PurangAbolmaesumiSeptimiuESalcudeanWen-hongZhuMohammadRezaSirouspour2002} have proposed a mechanism which achieves a high degree of probe manipulability and safety. The robot actuation has been moved to the base of the system, thus minimizing its size and weight. Other systems have been developed which separate the probe positioning into two stages: approximate probe placement and finer view-plane adjustments. The first can be achieved by a passive positioning mechanism, which is operated by a clinician, while the latter is obtained with an active end-effector. A system based on cables which are driven by McKibben actuators has been proposed \cite{VilchisGonzales2001}. The antagonistic configuration of the cables is employed to position the ultrasound probe on a patient. The system is tele-operated by a sonographer. Researchers from Waseda University first proposed this concept and corresponding design in \cite{Nakadate2009}, in which the end-effector is driven through a parallel mechanism. Similarly, a consortium of researchers have developed a system with active end-effector with the aim of remote tele-diagnosis \cite{Gourdon1999, Arbeille2003, Vieyres2003}. The system has since been trialled for remote scans \cite{Arbeille2005} and translated to a commercial product (MELODY, AdEchoTech, Naveil, France). Despite the scanning being performed remotely, the design of the system suggests, however, that the assisting operator is still required to apply the necessary force to maintain a stable contact.

Maintaining stable mechanical coupling between ultrasound probe and patient tissue is of paramount importance for ensuring a high-quality image. Approaches to achieve this involve controlling the contact force directly or establishing an elastic contact between the position-controlled device and the patient. While the first has been researched extensively \cite{Siciliano1999RobotControl}, \cite{Fang2017} and can be commonly found in various forms of industrial applications, the latter has found more attention in recent years due to an increased demand in cost effective force control and -limiting solutions for human robot collaboration tasks \cite{McMahan2006, Eiberger2008}. Series-elastic actuators have been developed to provide passive compliance in actuated robotic joints \cite{PrattSeriesActuators}. While providing a degree of compliance, this has the disadvantage that a collision or undesired contact in a direction other than the joint axis cannot be compensated for. We have trialled safety clutches for the use in ultrasound robots which exhibit compliant behaviour once disengaged through an excess force \cite{wang2019analysis, wang2019design}, \cite{Mathur2019}. This, however, renders the system uncontrollable and requires reengaging the clutch mechanism for further operation. In this work, we make use of an elastic soft robotic system, which is aimed at overcoming aforementioned limitations.
Soft robotics technologies have opened up new design paradigms for robotic systems through the use of elastic and deformable materials and structures \cite{Laschi2016, Polygerinos2017}. Soft robotics systems are commonly designed to interact with or conform to environmental contacts. This allows soft robotic manipulators to exhibit highly dexterous manoeuvrability in for example surgical \cite{Cianchetti2014, Marchese2014, Kahrs2015} or search and rescue operations \cite{Hawkes2017}. In these scenarios, however, soft robots are not applied to tasks which require significant loadbearing capabilities, predominantly due to their low stiffness. To bridge the trade-off between manoeuvrability and stiffness, research has been driven towards systems with variable stiffness capabilities. A comprehensive overview of stiffening technologies is given in \cite{Manti2016}. For applications in which softness is desired, high loadings are demanded and stiffening mechanisms are not suitable, soft robotic systems tend to be combined with external constraints to ensure structural integrity. This is commonly found in exoskeleton research and rehabilitation robotics. Examples include full body, soft exosuits \cite{Wehner2013AAssistance}, lower limb exoskeletons \cite{Costa2006} and hand exoskeletons for post-stroke rehabilitation \cite{Chiri2012, Stilli2018AirExGlovePatients}.

In our previous work, we identified the advantages of soft robotics technology in ultrasound interaction tasks compared to rigid state-of-the-art robots and showed an initial proof-of-concept of a parallel soft robotic end-effector with the right characteristics for medical ultrasound tasks \cite{Lindenroth2017}. We now derive a novel soft robotic end-effector which is capable of safely acquiring standard views in extracorporeal diagnostic foetal ultrasound (US). We select foetal US as an initial application  due to its high demands to robot safety. We evaluate the performance of our system with respect to derived specifications and show that the proposed system is capable of acquiring a set of standard view-plane required for the assessment of the foetus.  The robot utilizes linear soft fluidic actuators (SFAs) which are arranged in parallel around the ultrasound probe to provide high axial loadbearing capabilities and high lateral compliance, thus enabling adaptability and safety in the patient interaction. 
The individual contributions of this study are:
\begin{figure*}[t!]
    \centering
    \includegraphics[width=\linewidth]{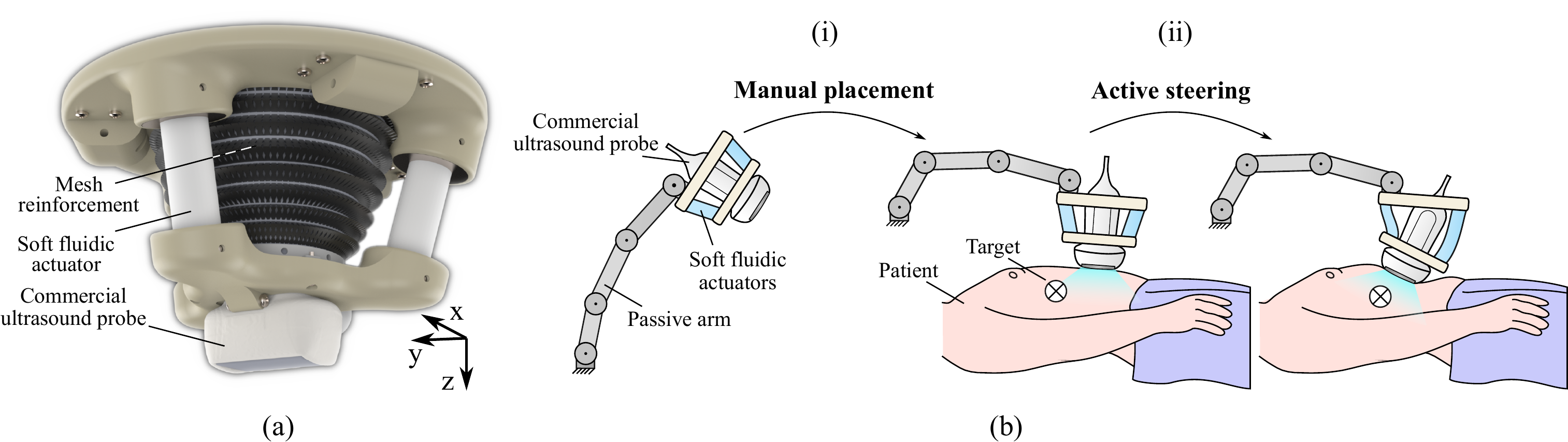}
    \caption{Proposed design of the soft robotic end-effector (a) and workflow (b) for obtaining a desired view through manual placement in the approximate region of interest (i) and active steering of the probe towards the desired view-plane (ii).}
    \label{fig:SEE_design}
\end{figure*}
\begin{itemize}
\item Clinical investigation to determine workspace and force requirements for view-plane adjustments in foetal diagnostic ultrasound imaging. 
\item Design and verification of a soft robotic end-effector which satisfies the derived clinical requirements in workspace and force. It employs robust linear soft fluidic actuators, for which a novel injection-based fabrication is derived, and undesired twist is prevented through a mesh constraint.
\item Definition and validation of a lumped stiffness model to describe the motion of the soft robotic end-effector in the absence and presence of external loading. 
\end{itemize}
The controllability and imaging capabilities of the integrated system are validated in position control and US phantom experiments respectively.

The paper is structured in the following way. In Section \ref{Design} the system requirements are determined, and the robot design is introduced. Based on the design of the system, Section \ref{Modelling} derives a kinetostatic model. Methodologies for the actuation and control of the system are presented in Section \ref{ActuationAndControl}. In Section \ref{Experiments} the mechanical properties of the system and its workspace are evaluated. Results are presented in section \ref{Results}. The proposed model is validated and the position controller performance, as well as the imaging capabilities of the system, are assessed.

\section{Methods}
Prenatal foetal ultrasound is a routine diagnostic procedure for pregnant women to determine birth defects and abnormalities in the foetus. Common checks include measuring the foetus’ biparietal diameter (BPD), its head and abdominal circumferences (HC and AC) as well as its femur length (FL) \cite{Salomon2011}.

In this work we focus on obtaining HC, AC and FL standard view-planes. We establish the clinical requirements to the contact force and movement range of the ultrasound probe for bespoke application and derive a suitable design for a soft robotic end-effector (SEE).

\begin{figure}[t]
    \centering
    \includegraphics[width=\linewidth]{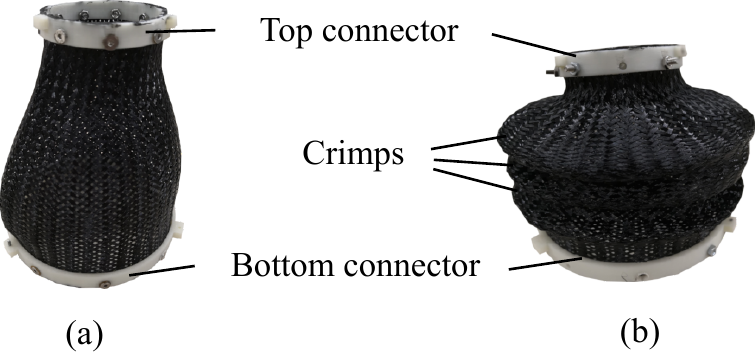}
    \caption{Braided nylon mesh uncrimped (a) and crimped (b).}
    \label{fig:Mesh}
\end{figure}

\subsection{Design}
\label{Design}
\subsubsection{Clinical data acquisition and processing}
Pregnant women between 18 to 24 weeks of gestation underwent research ultrasound scans at St Thomas’ Hospital (Study title: \emph{Intelligent Fetal Imaging and Diagnosis (iFIND)-2: Further Ultrasound and MR Imaging}, Study reference: 14/LO/1806). Trained sonographers performed the foetal ultrasound scan using a standard ultrasound probe (X6-1, Philips, Amsterdam, Netherlands) which is connected to an ultrasound scanner (EPIQ7, Philips, Amsterdam, Netherlands). The probe was placed in a holder as detailed in \cite{Noh2015}. This holder incorporated an electromagnetic (EM) tracking sensor (Aurora, NDI, Ontario, Canada) and six axis force-torque sensor (Nano 17, ATI, Apex, USA), which allowed measurements of the position and orientation of the probe, and the force applied at the probe face to be measured throughout the scan. The recorded tracking and force data of six patients were analysed by extracting time ranges during which standard fetal anomaly views were imaged. These included HC, AC and FL views. Each time range consisted of the few seconds when the sonographer had placed the probe in the correct anatomical region and was adjusting the probe to find the ideal view. For each view the tracking data were analysed to find the range of positions and orientations in the three axes separately. The X and Y axes show movement in the horizontal plane of the scanning bed (left to right on the patient, and foot to head, respectively), and the Z axis shows vertical movement. Orientation ranges are given in probe coordinates, with yaw showing axial rotation, pitch showing elevational tilting out of the image plane, and roll showing in-plane lateral tilting. Forces were analysed by dividing the measured force vector into normal and tangential components applied to the surface. The local surface angle was determined at each measurement by fitting an ellipsoidal shape to the tracking data of the scan. The 95th percentile of the forces measured within a time range gives an indication of the maximum force that must be applied by the probe.

\subsubsection{Mechanism requirements and synthesis}
Following the results of the clinical data analysis, it is found that the soft robotic end-effector must at least satisfy the following requirements
\begin{itemize}
    \item Be able to withstand mean axial and transversal contact forces of 8.01N and 4.42N without significant deterioration of the imaging view.
    \item Achieve an axial extension along Z of 5.22mm and transversal translations in X and Y of 7.75mm.
    \item Achieve rotations of 5.08$\degree$ around X and Y.
\end{itemize}

\begin{figure}
    \centering
    \includegraphics[width=0.9\linewidth]{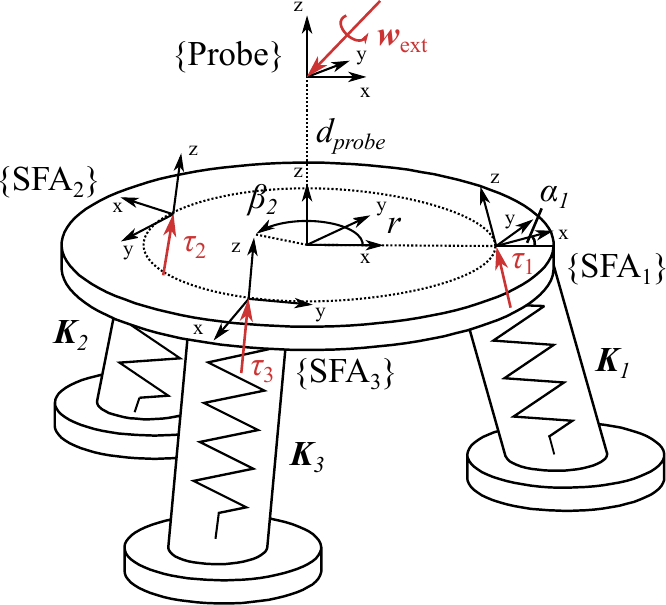}
    \caption{Free body diagram of SEE model definition}
    \label{fig:Model}
\end{figure}
To maintain a high degree of safety when interacting with the device, the SEE should furthermore comprise of a low transversal stiffness. This allows both the operating clinician and patient to manually displace the probe in case of discomfort.

As the investigated system is compliant, its deflection has to be considered when determining if a position is achievable. Taking into account normal and tangential forces applied during the scanning, the system must satisfy the following conditions
\[
    \boldsymbol{\delta}_{SEE} \geq \boldsymbol{\delta}_{req} +\boldsymbol{\delta}_{f} \quad \text{with} \quad \boldsymbol{\delta}_{f} = \boldsymbol{K}^{-1}_{min}\boldsymbol{f}_{req}
\]
Where $\boldsymbol{\delta}_{f}$ is a deformation induced by external forces, $\boldsymbol{f}_{req}$ is a vector of the required forces and $\boldsymbol{K}_{min}$ is the minimum system stiffness throughout the workspace. $\boldsymbol{\delta}_{req}$ and $\boldsymbol{\delta}_{SEE}$ are vectors of the required and achievable translations respectively. As only tip forces are considered in this work, tilting effects induced by external moments at the SEE tip are ignored and forces are assumed to only affect the tip position.
A soft robotic design based on soft fluidic actuators (SFAs), which have previously been presented in \cite{Lindenroth2017}, is proposed. It is comprised of two rigid platforms which serve as base and transducer holder respectively. The platforms are connected through a set of three soft fluidic actuators which are arranged in a parallel fashion at $120\degree$ intervals. To allow for sufficient space for the ultrasound transducer cable, the actuators are tilted at an angle of 15$\degree$. An overview of the design is shown in Fig. \ref{fig:SEE_design}a). Whilst a rigid mechanism of such configuration would be over-constrained and thus unable to move, the elasticity of the SFAs allows the SEE to perform bending (coupled translation and rotation) and axial extension motions.

As the SFAs are tilted, axial extension causes the SFAs to bend into an S-shaped configuration. This allows for the SEE to be axially compliant whilst exhibiting a high degree of load-bearing capabilities, which is further investigated in Section \ref{Stiffness}. Furthermore, curving into an S-shaped configuration eliminates the possibility of unstable buckling to occur in the SFAs, as shown in Section \ref{Results_Stiffness}.

A common problem in such a proposed soft robotic system is the low stiffness along its twist axis. To improve the stability of the system against twist deformations, a nylon fibre mesh is attached to base and transducer platforms, which acts as a mechanical constraint between the two. To reduce unwanted buckling behaviour, crimps can be added to the mesh by deforming and heat-treating it. Examples of uncrimped and crimped meshes are shown in Fig. \ref{fig:Mesh}. Thus, axial rotation of the ultrasound transducer is not considered in this study, as it could be added by simply applying a rotating mechanism to the base of the SEE, which would function as a stiff rotational axis in conjunction with the mesh constraint.

The workflow of imaging using the SEE is shown in Fig. \ref{fig:SEE_design}b). Once the SEE is manually placed in the approximate area of the target view using a passive positioning arm, it is fixed on the patient. The ultrasound probe is then actively steered either in a tele-operated manner by a sonographer or in an autonomous fashion using pose or image feedback. As the loadbearing is achieved by the SEE, contact forces the sonographer is required to apply are minimized, which presumably has an impact on the ergonomics of the sonographer.

\subsection{Kinetostatic modelling}
\label{Modelling}
To determine the ultrasound probe pose under internal fluid volume variation and external loading a kinetostatic model is derived according to \cite{Klimchik2018}. A free body diagram of the model is shown in Fig. \ref{fig:Model}. In the following, a vector denoted as $\boldsymbol{w}_f$ represents a 6 degree of freedom wrench in an arbitrary frame $f$  such that  $\boldsymbol{w}_f=[F_x^f,F_y^f,F_z^f,M_x^f,M_y^f,M_z^f ]^T$ with forces $\boldsymbol{F}$ and moments $\boldsymbol{M}$. Similarly, $\boldsymbol{\tau}_f$ denotes a reaction wrench in the local SFA frame, which is of the same form as $\boldsymbol{w}_f$. Vectors noted as $\delta x_f$ indicate infinitesimally small displacements in frame f of the form $\delta \boldsymbol{x}_f=[u_x^f,u_y^f,u_z^f,v_x^f,v_y^f,v_z^f ]^T$ with translations $u$ and rotations $v$.

Let $\boldsymbol{w}_{ext}$ be a vector of forces and moments applied to the tip of the ultrasound transducer. Under static equilibrium conditions, the following holds for a single actuator

\begin{equation}\label{eq:StatEq}
\boldsymbol{w}_{ext}=\boldsymbol{w}_\theta + \boldsymbol{w}_V
\end{equation}

Where $\boldsymbol{w}_\theta$ is the wrench caused by the elastic deformation of the SFA and $\boldsymbol{w}_V$ is the reaction wrench caused by the constrained hydraulic chamber. Both are expressed in the tip frame of the system.
The tip wrenches $\boldsymbol{w}_\theta$ and $\boldsymbol{w}_V$ can be expressed relative to their local frames by
\begin{equation}\label{eq:Ad}
\begin{split}
\boldsymbol{w}_\theta & =\boldsymbol{J}_\theta(\boldsymbol{x}) \boldsymbol{\tau}_\theta \\ \boldsymbol{w}_V & =\boldsymbol{J}_V(\boldsymbol{x})\tau_V
\end{split}
\end{equation}
Where $\boldsymbol{\tau}_\theta$ is a vector of local reaction forces and moments caused by the SFA deformation and $\boldsymbol{\tau}_V$ is the uniaxial reaction force of the volumetric constraint in the actuator. The matrices $\boldsymbol{J}_\theta(\boldsymbol{x})$ and $\boldsymbol{J}_V(\boldsymbol{x})$ are defined by
\begin{equation}\label{eq:Ad}
\begin{aligned}
\boldsymbol{J}_\theta(\boldsymbol{x}) &= 
\begin{bmatrix}
    \boldsymbol{R}(\boldsymbol{x})    & \boldsymbol{0}\\
    \boldsymbol{0} & \boldsymbol{R}(\boldsymbol{x})
\end{bmatrix}\boldsymbol{Ad} \\
\boldsymbol{J}_V(\boldsymbol{x}) &= 
\begin{bmatrix}
    \boldsymbol{R}(\boldsymbol{x})    & \boldsymbol{0}\\
    \boldsymbol{0} & \boldsymbol{R}(\boldsymbol{x})
\end{bmatrix}\boldsymbol{Ad}_z
=
\begin{bmatrix}
    \boldsymbol{R}(\boldsymbol{x})    & \boldsymbol{0}\\
    \boldsymbol{0} & \boldsymbol{R}(\boldsymbol{x})
\end{bmatrix}\boldsymbol{\hat{H}}
\end{aligned}
\end{equation}
$\boldsymbol{R}(\boldsymbol{x})$ is the rotation matrix of the current tip deflection. Matrix $\boldsymbol{Ad}$ is the wrench transformation matrix relating the local SFA frame to the tip frame by 
\begin{equation}
\boldsymbol{Ad} = 
\begin{bmatrix}
    \boldsymbol{R}_0    & \boldsymbol{0} \\
    \boldsymbol{D}_0\boldsymbol{R}_0 & \boldsymbol{R}_0
\end{bmatrix}
\end{equation}
Where $\boldsymbol{R}_0$ is the spatial rotation of the respective frame and $\boldsymbol{D}_0$ is the cross-product matrix with the translation vector $\boldsymbol{d}_0 = [d_x, d_y, d_z]$. $\boldsymbol{\hat{H}}$ is for a single SFA a 6x1 vector containing the third column of $\boldsymbol{Ad}$.

Considering the elastic behaviour of the SFA, its reaction force $\boldsymbol{\tau}_\theta$ caused by an infinitesimally small, local displacement $\delta\boldsymbol{x}_\theta$ can be written as
\begin{equation}\label{eq:Hook}
\boldsymbol{\tau}_\theta=\boldsymbol{K}_\theta \delta \boldsymbol{x}_\theta
\end{equation}

Where the SFA stiffness $\boldsymbol{K}_\theta$ is defined as a Timoshenko beam element with
\[\label{xx}
\boldsymbol{K}_\theta =
\begin{bmatrix}
    \frac{12EI}{(1+\Phi)L^3}    & 0     & 0     & 0     & \frac{6EI}{(1+\Phi)L^2}   & 0\\
    0       & \frac{12EI}{(1+\Phi)L^3}  & 0     & \frac{-6EI}{(1+\Phi)L^2}  & 0     & 0 \\
    0       & 0                & \frac{EA}{L}    & 0     & 0     & 0\\
    0       & \frac{-6EI}{(1+\Phi)L^2}  & 0             & \frac{(4+\Phi)EI}{(1+\Phi)L}  & 0     & 0\\
    \frac{6EI}{(1+\Phi)L^2}  & 0 & 0 & 0             & \frac{(4+\Phi)EI}{(1+\Phi)L}  & 0\\
    0       & 0                 & 0     & 0   & 0     & \frac{GJ}{L}
\end{bmatrix}
\]

$L$ describes the length of the SFA, $A$ it’s cross-sectional area, $E$ its Young’s modulus, $I$ the area moment of inertia, $G$ its shear modulus and $J$ the torsion constant. The Timoshenko coefficient $\Phi$ is defined as 
\[
\Phi=\frac{12EI}{\frac{A}{\alpha}GL^3}
\]
with the Timoshenko coefficient $\alpha$. An overview of the SFA constants is given in Table \ref{tab:ModelParameters}.

\begin{table}[h!]
\centering
\caption{SFA model parameters}
\label{tab:ModelParameters}
\setlength\tabcolsep{2pt}
\begin{tabular}{lll}
\toprule
Constant & Value & Description\\ \midrule
 $L$ & $45$mm & Initial length\\
 $A$ & $\pi \cdot 10^2\text{mm}^2$ & Cross-sectional area\\
 $a$ & $\pi \cdot 6.9^2 \text{mm}^2$* & Fluid channel area\\
 $E$ & $301.51$kPa ** & Young's modulus\\
 $I$ &  $1200\text{cm}^4$ ** & Area moment of interia\\
 $G$ &  $0.5$E & Shear modulus\\
 $J$ &  $0.5\pi\cdot10^4\text{mm}^4$ & Torsion constant\\
 $\alpha$ & $5/6 $ & Timoshenko coefficient\\ \bottomrule
 \multicolumn{3}{l}{* obtained in Section \ref{SFA_characterization}; ** obtained in Section \ref{Model_validation}}
\end{tabular}
\end{table}

Whilst parameters $L$ and $A$ are obtained from the SFA geometry, the torsion constant of a beam with circular cross-section can be expressed as $J = 0.5\pi r^4$ and its Timoshenko coefficient is defined as $5/6$ \cite{Matrix}. The shear modulus $G$ is approximated as half the Young's Modulus.

For a given SFA volume, the kinematic relationship between an infinitesimal small volume change $\delta V$ of the SFA and the displacement of the ultrasound tip frame is given by

\begin{equation}\label{eq:Constraint}
\delta V/a=\boldsymbol{J}_V^T \delta\boldsymbol{x}_{tip}
\end{equation}
Where $a$ is the cross-sectional area of the fluid actuation channel. The kinematic motion of the tip frame caused by the SFA deflection can be defined as
\begin{equation}\label{eq:Kin}
\delta\boldsymbol{x}_{\theta}=\boldsymbol{J}_\theta^T \delta \boldsymbol{x}_{tip}
\end{equation}
Substituting Equation \ref{eq:Kin} into \ref{eq:Hook} yields
\begin{equation}\label{eq:DeflForce}
\boldsymbol{\tau}_\theta=\boldsymbol{K}_\theta  \boldsymbol{J}_\theta^T \delta \boldsymbol{x}_{tip}
\end{equation}

Applying Equations \ref{eq:Ad} and \ref{eq:DeflForce}, the static equilibrium condition in Equation \ref{eq:StatEq} can be written as
\begin{equation}\label{eq:StatEqFinal}
\boldsymbol{w}_{ext}=\boldsymbol{J}_\theta\boldsymbol{K}_\theta\boldsymbol{J}^T_\theta\delta\boldsymbol{x}_{tip} + \boldsymbol{J}_V\tau_V
\end{equation}
Equation \ref{eq:StatEqFinal} can be combined with the imposed kinematic constraint defined by Equation \ref{eq:Constraint} to a linear equation system of the form

\begin{equation}\label{eq:EqSys1}
\begin{bmatrix}
    \boldsymbol{w}_{ext}\\
    \delta V/a
\end{bmatrix} =
\begin{bmatrix}
    \boldsymbol{J}_\theta\boldsymbol{K}_\theta\boldsymbol{J}_\theta^T & \boldsymbol{J}_V\\
    \boldsymbol{J}_V^T & \boldsymbol{0}
\end{bmatrix}
\begin{bmatrix}
    \delta \boldsymbol{x}_{tip}\\
    \tau_V
\end{bmatrix}
\end{equation}

The deflection of the ultrasound transducer tip and internal reaction of the system can consequently be found through matrix inversion

\begin{equation}\label{eq:EqSys2}
\begin{bmatrix}
    \delta \boldsymbol{x}_{tip}\\
    \tau_V
\end{bmatrix} =
\begin{bmatrix}
    \boldsymbol{J}_\theta\boldsymbol{K}_\theta\boldsymbol{J}_\theta^T & \boldsymbol{J}_V\\
    \boldsymbol{J}_V^T & \boldsymbol{0}
\end{bmatrix}^{-1}
\begin{bmatrix}
    \boldsymbol{w}_{ext}\\
    \delta V/a
\end{bmatrix}
\end{equation}
The formulation can be expanded to a number of $n$ SFAs by considering a lumped stiffness $\boldsymbol{K}$ in the probe tip frame. As the actuators are aligned in a parallel configuration, it can be defined by

\begin{equation}\label{eq:Lump}
\boldsymbol{K} = \sum_{i=1}^{n}\boldsymbol{J}_\theta^i \boldsymbol{K}_\theta^i {\boldsymbol{J}_\theta^i}^T
\end{equation}
The matrix $\boldsymbol{J}_V$ is adopted by appending the respective columns of the wrench transformation matrix of actuator $i$ $\boldsymbol{Ad}_z^i$ to $\boldsymbol{\hat{H}}$ such that
\begin{equation}\label{eq:J_V}
^n\boldsymbol{J}_V = \begin{bmatrix}
    \boldsymbol{R}(\boldsymbol{x})    & \boldsymbol{0}\\
    \boldsymbol{0} & \boldsymbol{R}(\boldsymbol{x})
\end{bmatrix}
[\boldsymbol{Ad}_z^1, \boldsymbol{Ad}_z^2, ..., \boldsymbol{Ad}_z^n]
\end{equation}
The kinematic constraint relationship then becomes
\begin{equation}\label{eq:KinVector}
\delta \boldsymbol{V}/a = ^n\!\boldsymbol{J}_V^T \boldsymbol{x}_{tip}
\end{equation}
Where $\delta \boldsymbol{V}$ is an $n \times 1$ vector of SFA volume changes. Consequently, $\tau_V$ is expanded to an $n \times 1$ vector containing $n$ local reactions in the form $\boldsymbol{\tau}_V=[\tau_{V,1},\tau_{V,2}...,\tau_{V,n}]^T$.

To account for changes in matrices $\boldsymbol{J}_\theta$ and $\boldsymbol{J}_V$ for a given motion, the model is solved numerically by dividing the applied external wrench and induced volume vectors into small increments $[\Delta \boldsymbol{w}_{ext}, \Delta \boldsymbol{V}]^T$. After each iteration, $\boldsymbol{R}(\boldsymbol{x})$ is updated according to the previous tip pose.

\begin{figure}[!t]
\centering
\includegraphics[width=\linewidth]{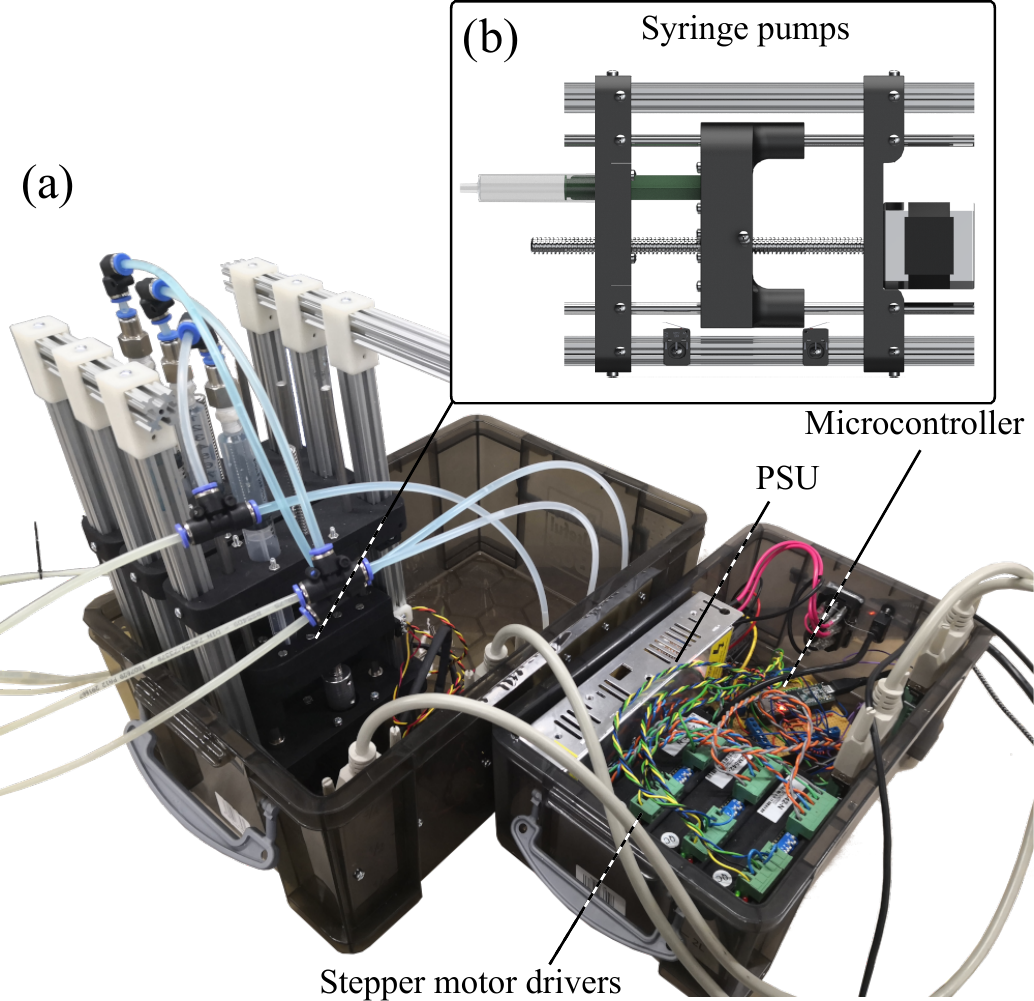}
\caption{Actuation unit (a) with syringe pumps (b)  and controller system}
\label{fig:SP}
\end{figure}

For the given number of three SFAs, the update rule for the numerical solution is defined by
\begin{equation}\label{eq:EqSys2}
\begin{bmatrix}
    \boldsymbol{x}^{k}_{tip}\\
    \boldsymbol{\tau}_V^{k}
\end{bmatrix} =
\begin{bmatrix}
    \boldsymbol{x}^{k+1}_{tip}\\
    \boldsymbol{\tau}_V^{k+1}
\end{bmatrix} + 
\begin{bmatrix}
    \boldsymbol{K} & ^3\boldsymbol{J}^k_V\\
    ^3{\boldsymbol{J}^k_V}^T & \boldsymbol{0}
\end{bmatrix}^{-1}
\begin{bmatrix}
    \Delta \boldsymbol{w}_{ext}\\
    \Delta \boldsymbol{V}/a
\end{bmatrix}
\end{equation}
For iteration step $k$.

\subsection{Actuation and control}
\label{ActuationAndControl}
The SEE is actuated by inflating respective SFAs with a working fluid. As shown in our previous work \cite{Lindenroth2016}, we utilize custom hydraulic syringe pumps (Fig. \ref{fig:SP}b)) which are driven by stepper motors (Nema 17, Pololu Corporation, Las Vegas, USA) to induce volume changes in the SFAs. The pumps are controlled with a microcontroller (Teensy 3.5, PJRC, Sherwood, USA) which communicates via a serial interface with a PC running ROS (Intel Core I7-7700HQ, XPS15 9560, Dell, Texas, USA). The PC generates demand velocities or positions for the microcontroller and solves the previously-defined kinetostatic model to determine the system Jacobian for a given pose. Furthermore, the laptop handles interfaces with peripherals such as a joystick for teleoperation (Spacemouse Compact, 3dconnexion, Monaco) and an electromagnetic tracking (EM) system for closed-loop position control (Aurora, NDI, Ontario, Canada).

\begin{figure}[t!]
    \centering
    \includegraphics[width=\linewidth]{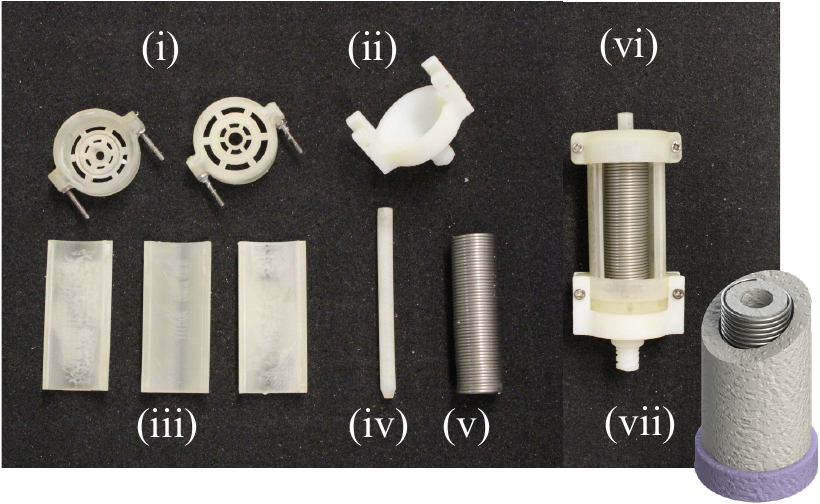}
    \caption{Overview of mould components (i)-(vi) and drawing of final SFA (vii)}
    \label{fig:Mould}
\end{figure}

The linear soft fluidic actuators which are utilized to drive the system have first been conceptualized in our previous work \cite{Lindenroth2017}. They are comprised of a silicone rubber body (Dragonskin 10-NV, SmoothOn Inc, Pennsylvania, USA) and stiffer silicone rubber endcaps (SmoothSil 945, SmoothOn Inc, Pennsylvania, USA). A helical constraint is inserted into the silicone to counteract radial expansion of the actuator upon inflation. This, in combination with the stiff endcaps, allows for the actuators to maintain its form and only expand in the direction of actuation. The moulding process of creating SFAs has been significantly improved from our previous work. For the radial constraint an extension spring (Fig. \ref{fig:Mould}(v)) is used. The liquid silicone rubber is injected through an inlet (Fig. \ref{fig:Mould}(ii)) using a syringe instead of being poured into the mould. This has the significant advantage for the user to be able to pre-assemble the mould without having to manually wind the constraint helix, as it has been commonly done in soft fluidic actuators \cite{Suzumori2002}. In combination with the injection of the silicone this could reduce variations in the fabrication process. A drawing of a finished actuator is shown in Fig. \ref{fig:Mould}(vii). The combination of radial constraint and stiff endcaps allows for the actuators to be driven efficiently with a volumetric input without exhibiting a nonlinear relationship between input volume and output length change due to bulging, which is investigated in Section \ref{SFA_characterization}.

In this work, two methods for controlling the ultrasound probe pose are investigated. A joystick-based teleoperated open-loop controller is implemented to allow a sonographer to steer the probe according to the acquired ultrasound image stream. For this purpose, the aforementioned joystick is used. The axial motion of the joystick is linked to a translation of the SEE in Z-direction while the two tilt axes of the joystick are mapped to the X- and Y-rotation axes of the SEE. The high-level controller generates syringe pump velocities according to
\begin{equation}
    \boldsymbol{\dot{V}_d} = \boldsymbol{J}_V^T  \boldsymbol{v_{cart}}
\end{equation}
Where $\boldsymbol{\dot{V}}_d$ is the desired SFA velocity, $\boldsymbol{v_{cart}}$ the target velocity in Cartesian space and $\boldsymbol{J_V}^T$ the actuation matrix of the system which has been derived in Section \ref{Modelling}.

\textcolor{black}{
A closed-loop controller is integrated to drive the ultrasound probe tip position according to EM tracker feedback. For this purpose, a high-level trajectory generator continuously updates the demand position for the position controller, which generates in return demand volumes for the three syringe pumps according to the control law
\begin{equation}
    \boldsymbol{\Delta{V}_d} = \boldsymbol{J}_V^T \boldsymbol{U}
\end{equation}
Where $\Delta\boldsymbol{V}_d$ is the desired change in volume and $\boldsymbol{U}$ the control signal. A linear PI controller of the form 
\begin{equation}
    \boldsymbol{U} = \boldsymbol{K}_P\boldsymbol{X}_e + \boldsymbol{K}_I\int\boldsymbol{X}_e dt
\end{equation}
is employed, where $\boldsymbol{X}_e = \boldsymbol{X}_d - \boldsymbol{X}_c$. $\boldsymbol{X}_d$ and $\boldsymbol{X}_c$ are demanded and measured probe tip position respectively.}
\textcolor{black}{The gain matrices $\boldsymbol{K}_P=diag(k_P,k_P,k_P)$ and $\boldsymbol{K}_I=diag(k_I,k_I,k_I)$ contain the gain constants $k_P$ and $k_I$, which have been verified experimentally and are defined as $0.3 \frac{\text{ml}}{\text{mm}}$ and $0.03 \frac{\text{ml}\cdot s}{\text{mm}}$ respectively.}
The target points are generated at 2Hz while both the position controller and the kinetostatic model are updated at 30Hz. The low-level step generation for driving the syringe pumps is achieved with an update rate of 6kHz.

\section{\textcolor{black}{Experimental validation}}
\label{Experiments}

\subsection{SFA characterization}

Using the  \textcolor{black}{three SFAs to control the SEE pose} in an open-loop configuration requires the volume-extension relation to be predictable for any given point in time. \textcolor{black}{From the radial mechanical constraint incorporated in the SFA design it is assumed that the relationship between induced volume and SFA length change is linear. To verify this, the extension behaviour of a single SFA is investigated for different working fluid changes} using a linear rail setup. The position of the tip of the actuator is equipped with a slider and tracked using a linear potentiometer. Contact friction between the linear bearings and rails is minimized using lubrication and friction forces are therefore neglected in the evaluation of the results. Volume and extension data are tracked and synchronized using ROS.

\subsection{Stiffness characterization}
\label{Stiffness}
\begin{figure}[t!]
    \centering
    \includegraphics[width=\linewidth]{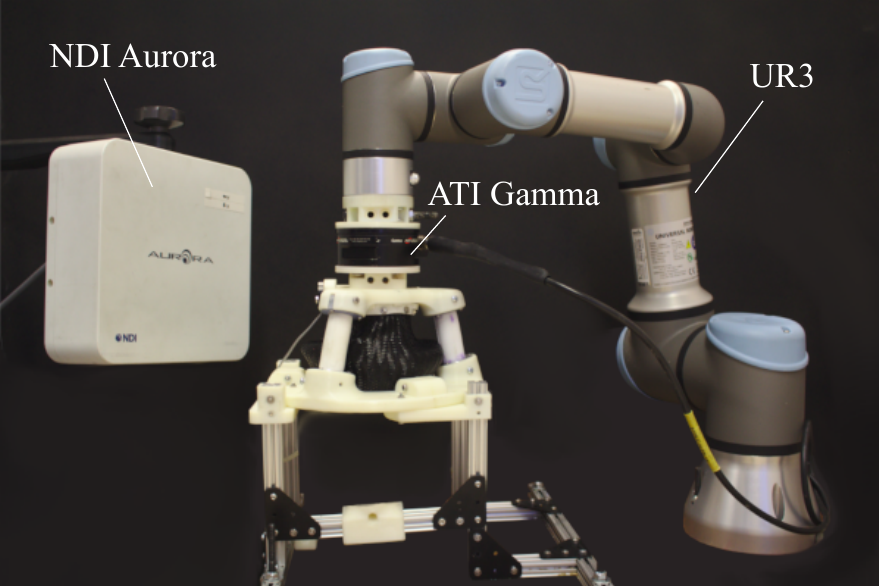}
    \caption{Experimental setup for stiffness characterization}
    \label{fig:UR3}
\end{figure}

\textcolor{black}{As the SEE is highly compliant, knowledge of its deformability under external loads is required to determine its efficacy to the given task. To verify the structural behaviour of the SEE under contact forces required for the clinical application, the stiffness of the system is characterized} with the setup shown in Fig. \ref{fig:UR3}. The SEE is mounted to a base plate and its tip is connected through a force-torque sensor (Gamma, ATI, Apex, USA) to a robot manipulator (UR3, Universal Robots, Odense, Denmark). To determine the stiffness of the SEE in a given direction, the manipulator moves the SEE in said direction and the resulting reaction force is measured\textcolor{black}{. The robot allows for an accurate, repeatable displacement of the SEE in a defined direction, thus isolating the desired DOFs. The payload of the system is with 3kg sufficiently high to withstand the induced reaction forces caused by the elastic deformation of the SEE}. The motions are repeated 10 times for each configuration. The linearized relationship between reaction force and manipulator displacement corresponds to the stiffness of the SEE.

The mesh reinforcement’s effect on the axial twist stiffness is determined by twisting the SEE repeatedly by $10\degree$ and measuring the z-axis moment. This is done for a configuration without mesh reinforcement, mesh reinforcement without crimps and mesh reinforcement with crimps.

\begin{figure}[t!]
    \centering
    \includegraphics[width=\linewidth]{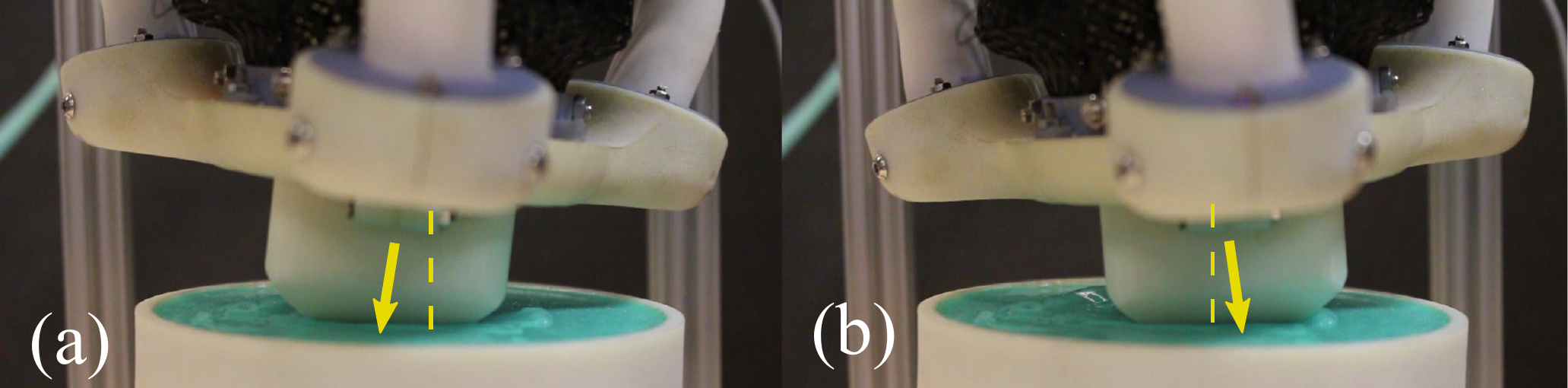}
    \caption{SEE moving in contact with soft rubber patch. \textcolor{black}{The tip pose change with respect to the SEEs origin is highlighted with an arrow.}}
    \label{fig:ContactPatch}
\end{figure}
The directional lateral stiffness is obtained by displacing the SEE tip radially in a defined direction over a distance of 10mm. This is repeated for four inflation levels (25\%, 50\%, 75\% and 100\% of the maximum SFA volume) and for directions between 0$\degree$ and $345\degree$ in $15\degree$ increments around the z-axis. The axial stiffness which corresponds to each extension is determined by displacing the SEE tip in negative z-direction by 1.5mm for 25\% and 50\% inflation, and by 2.5mm for 75\% and 100\% extension.

\subsection{Workspace and repeatability}
\label{Experimental_WS}
\textcolor{black}{To verify whether the attainable motions of the SEE satisfy the imposed clinical requirements for the ultrasound probe motion, the workspace of the SEE is mapped for achievable volumetric inputs.} The \textcolor{black}{SEE pose} is measured using an electromagnetic tracker (6DOF Reference, Aurora, NDI, Ontario, Canada) which is attached to the side of the SEE tip. The pose of the ultrasound probe tip is calculated with the known homogeneous transformation between tracker and tip. The SFA volumes are varied between 0\% and 100\% in 10\% increments and the resulting static tip pose is determined with respect to its deflated state.

The repeatability in positioning the tip of the SEE is determined by repeatedly approaching defined SFA volume states and measuring the tip pose. A set of 6 states is defined and the resultant trajectory is executed 50 times.

\subsection{Model validation}

The derived model is validated by comparing the workspace and corresponding SFA volumes to the calculated tip pose of the SEE. \textcolor{black}{For this purpose tip poses are calculated for each configuration achieved in Section \ref{Experimental_WS} and the error between model and measurement is determined.}

\subsection{Indentation behaviour}
\textcolor{black}{Whilst the abdomen exhibits an increased stiffness with the duration of the pregnancy and thus counteracts indentation of the ultrasound probe, deep tissue indentation in the early weeks can affect the positioning behaviour of the SEE.} To verify the effect a soft tissue-like contact has on the SEE, a soft mechanical phantom is created. The cylindrical phantom is moulded from a layer of Ecoflex Gel and a structural layer of Ecoflex 00-30 (SmoothOn Inc, Pennsylvania, USA).

The tip of the SEE is controlled to perform a line trajectory from its negative to positive x-axis limits at 60\% inflation. The tip pose is monitored with a magnetic tracker and contact forces between SEE and phantom are measured using aforementioned force sensor at the base of the phantom. The manipulator is used to test for different indentation depths from 0mm to 15mm in 5mm increments.

\begin{figure}[t!]
    \centering
    \includegraphics[width=\linewidth]{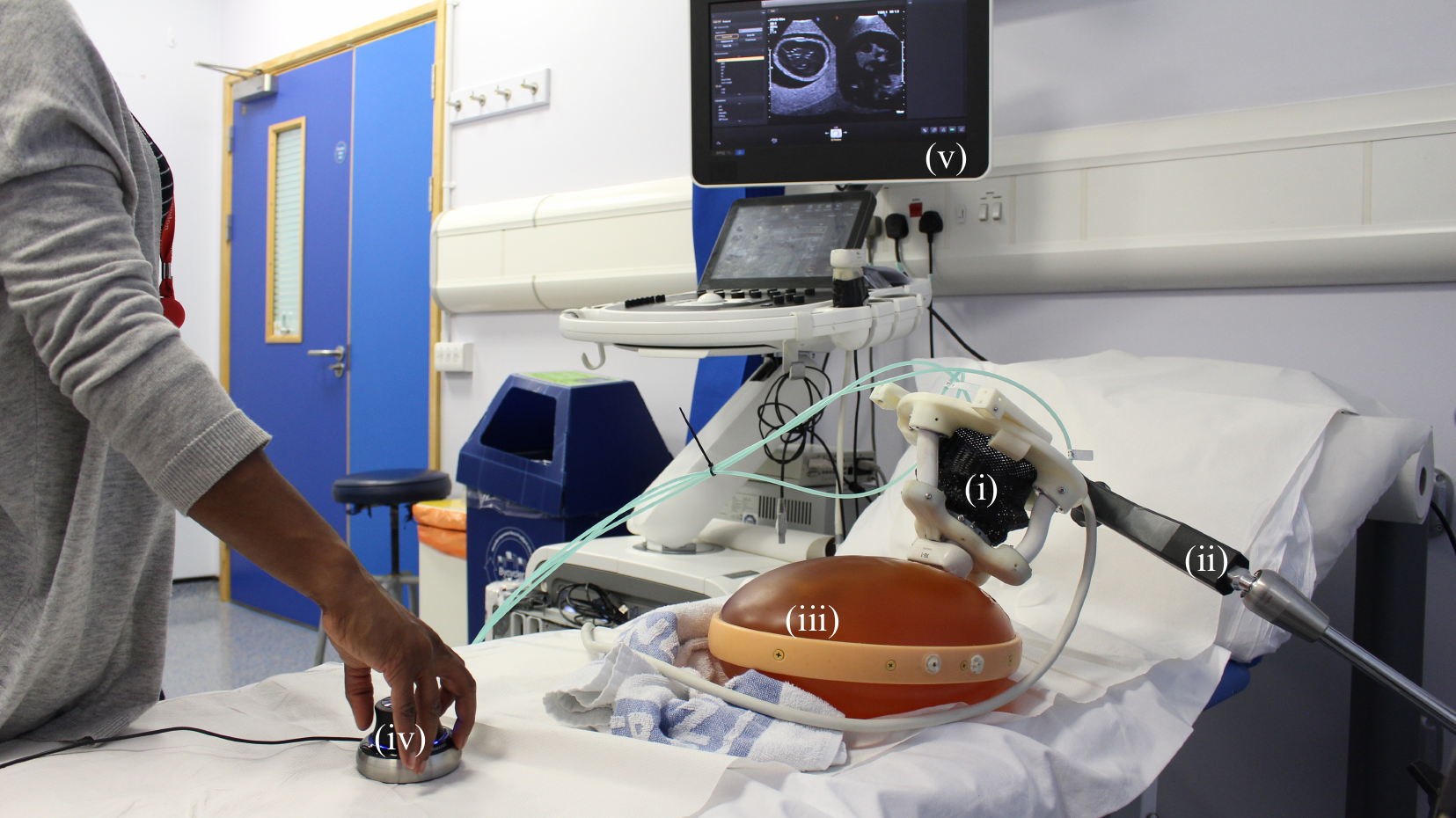}
    \caption{Sonographer performing SEE-assisted ultrasound scanning of a prenatal abdominal phantom (iii). The SEE (i) is attached to a passive arm (ii) and manually placed on the phantom. A joystick (iv) is used to manipulate the ultrasound probe under visual guidance of the acquired image (v).}
    \label{fig:Imaging}
\end{figure}

\subsection{Controllability}
\textcolor{black}{To achieve a desired view-plane in the ultrasound image, the probe attached to the SEE needs to be steerable accurately across the patient's body.} The controllability of the SEE is verified with the closed-loop position control system described in Section \ref{ActuationAndControl}. Target trajectories are defined as isosceles triangles with a base of 12.33mm and height of 10mm. For the tilted trajectory, the triangle is titled about one of its sides by 19$\degree$. The trajectory is tested in a planar and tilted configuration and tracked 3 times each.

To determine the controllability under an external load, a stiff silicone rubber patch is created as shown in Fig. \textcolor{black}{\ref{fig:ContactPatch}}. The patch is lubricated and positioned with its center at the tip of the SEE. To ensure contact with the patch, an initial axial force of 5N is generated by displacing the patch and running the position controller. This is repeated for planar and tilted configurations, where each trajectory is tracked 3 times.

\subsection{Sonographer-guided teleoperation}
The imaging capabilities of an ultrasound transducer guided by the SEE are verified using a prenatal abdominal phantom (SPACE FAN-ST, Kyoto Kagaku, Japan). The SEE is equipped with an abdominal ultrasound probe (X6-1, Philips, Amsterdam, Netherlands) which is connected to an ultrasound scanner (EPIQ7, Philips, Amsterdam, Netherlands). A passive positioning arm (Field Generator Mounting Arm, NDI, Ontario, Canada) is used to manually position the SEE in the region of interest on the phantom. The sonographer uses the provided ultrasound image feedback to steer the SEE with the connected joystick towards a desired view-plane.
The target view-planes are manually acquired using a handheld ultrasound probe. An overview of the experimental setup is shown in Fig. \ref{fig:Imaging}.

\section{Results}
\label{Results}

\begin{figure}[t!]
    \centering
    \includegraphics[width=\linewidth]{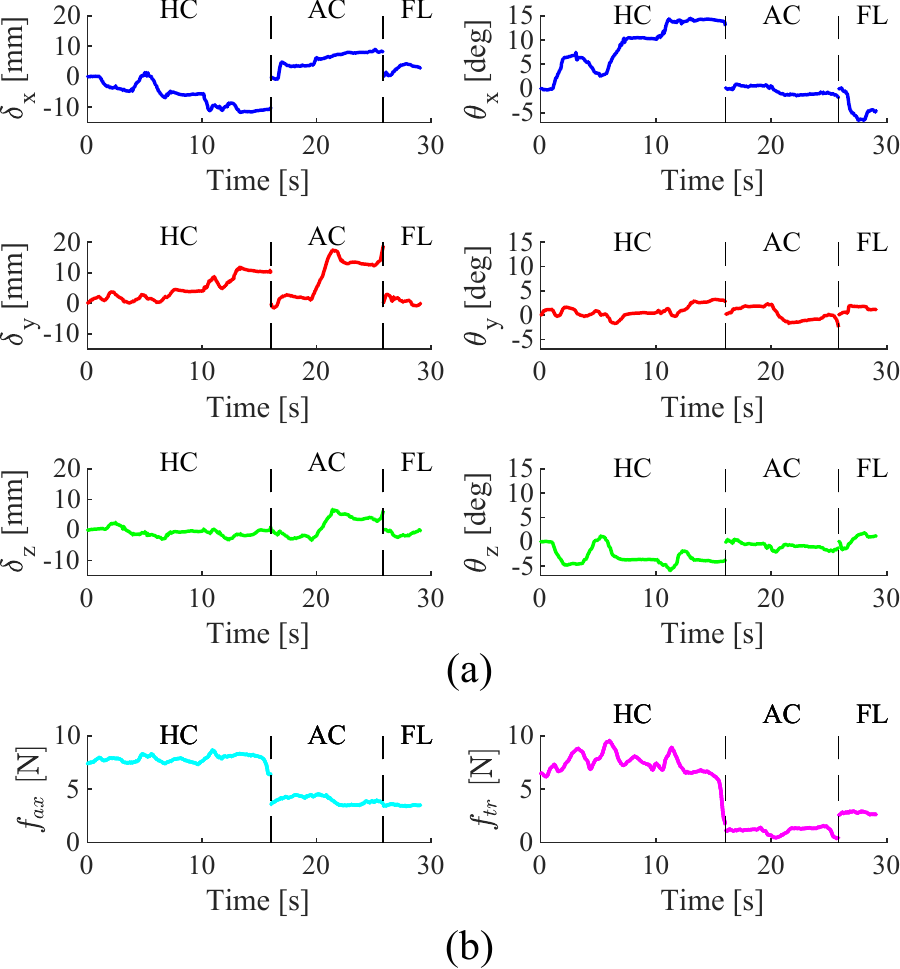}
    \caption{\textcolor{black}{Time series of probe pose (a) and tip force (b) for subject 5. Data between motions corresponding to standard views HC, AC and FL have been omitted.}}
    \label{fig:ClinicalData}
\end{figure}
\subsection{Clinical data}
\label{ClinicalData}
\textcolor{black}{The results of the clinical data acquisition are presented in Table \ref{tab:ClinicalData}. For each subject the maximum observed motion range in translation and rotation of the ultrasound probe is presented for the HC, AC and FL standard views. The presented forces correspond to the 95th percentile of the occurring normal and tangential force magnitudes. A time series of the probe pose and force data obtained for subject 5 is shown in Fig. \ref{fig:ClinicalData}.}

For subject 2 only HC and AC views were obtained. Translations and rotations are shown with respect to the patient bed. The normal force is assumed to be acting only in negative probe direction and the tangential force shows the vector magnitude of the tangential forces in X and Y.

\textcolor{black}{To obtain workspace requirements which are compatible with the obtained forces, it is divided into transversal and axial movements and transversal rotations. In this study, axial rotations of the probe are ignored. Workspace requirements for the SEE are consequently obtained by selecting the larger translation between X and Y for the transversal $\delta_{req}^{tr}$ and the translation in Z for the axial motion $\delta_{req}^{ax}$, thus resulting in a required cylindrical workspace of radius $\delta_{req}^{tr}$ and height $\delta_{req}^{ax}$. For the orientation, the required rotation is defined by $\theta_{req}^{tr}$. The mean required workspace from the clinical data is therefore
\begin{equation}
    \begin{split}
        \boldsymbol{\delta}_{req} = &[\delta_{req}^{ax}, \delta_{req}^{tr}]^T = [5.22\text{mm}, 7.75\text{mm}]^T\\
        \theta_{req} = &5.08\degree
    \end{split}
\end{equation}}
Corresponding maximum tilts of pitch and roll are in ranges of $\pm9.8\degree$ and $\pm12.9\degree$. The maximum occurring normal and tangential forces are 20.77N and $\pm$10.67N respectively.

\begin{table}[htbp]
  \centering
  \caption{Range of motion and contact force required to obtain a desired view in foetal ultrasound. Values used to generate the required SEE workspace are marked in blue.}
  \setlength\tabcolsep{2pt}
  \resizebox{\linewidth}{!}{ 
    \begin{tabular}{rccccccccc}
    \multicolumn{1}{c}{\multirow{2}[3]{*}{Subj.}} & \multirow{2}[3]{*}{View} & \multicolumn{3}{c}{Max. translation [mm]} & \multicolumn{3}{c}{Max. rotation [deg]} & \multicolumn{2}{c}{Force range [N]} \\
\cmidrule{3-10}      &   & X & Y & Z & Yaw & Pitch & Roll & Normal & Tangential \\
    \midrule
    \multicolumn{1}{c}{\multirow{3}[2]{*}{1}} & HC & \textbf{8.50} & 3.41 & \textbf{5.67} & \textbf{6.19} & \textbf{4.98} & \textbf{7.72} & \textbf{13.81} & 1.92 \\
      & AC & 4.49 & 5.60 & 2.76 & 3.44 & 3.45 & 3.17 & 4.10 & 2.60 \\
      & FL & 6.03 & \textbf{7.86} & 4.51 & 5.51 & 4.84 & 2.72 & 7.44 & \textbf{4.15} \\
    \midrule
    \multicolumn{1}{c}{\multirow{3}[2]{*}{2}} & HC & 4.95 & \textbf{6.45} & 2.96 & 5.74 & \textbf{5.91} & 10.09 & \textbf{14.09} & \textbf{6.15} \\
      & AC & \textbf{13.53} & 6.33 & \textbf{7.53} & \textbf{10.80} & 5.88 & \textbf{12.90} & 8.73 & 1.78 \\
      & FL & - & - & - & - & - & - & - & - \\
    \midrule
    \multicolumn{1}{c}{\multirow{3}[2]{*}{3}} & HC & \textbf{9.73} & \textbf{12.41} & 6.52 & 7.26 & \textbf{9.80} & 4.92 & \textbf{13.27} & \textbf{4.87} \\
      & AC & 1.53 & 9.37 & 2.60 & 4.23 & 3.62 & 4.10 & 5.55 & 1.40 \\
      & FL & 5.81 & 6.93 & \textbf{7.96} & \textbf{10.34} & 3.80 & \textbf{7.78} & 6.61 & 2.36 \\
    \midrule
    \multicolumn{1}{c}{\multirow{3}[2]{*}{4}} & HC & \textbf{11.87} & \textbf{8.36} & \textbf{7.02} & \textbf{14.84} & \textbf{4.70} & \textbf{9.39} & 3.47 & \textbf{3.62} \\
      & AC & 2.76 & 2.31 & 3.19 & 0.67 & 1.04 & 1.09 & \textbf{4.36} & 3.59 \\
      & FL & 4.64 & 4.51 & 4.64 & 1.82 & 3.55 & 2.07 & 3.61 & 2.27 \\
    \midrule
    \multicolumn{1}{c}{\multirow{3}[2]{*}{5}} & HC & \textbf{13.08} & 11.82 & 5.79 & \textbf{14.78} & \textbf{4.96} & \textbf{7.14} & \textbf{8.30} & \textbf{8.94} \\
      & AC & 9.77 & \textbf{19.91} & \textbf{10.22} & 2.83 & 4.55 & 2.65 & 4.46 & 1.49 \\
      & FL & 4.11 & 3.77 & 2.61 & 6.75 & 1.92 & 3.48 & 3.59 & 2.92 \\
    \midrule
    \multicolumn{1}{c}{\multirow{3}[2]{*}{6}} & HC & 2.49 & \textbf{10.18} & \textbf{7.55} & \textbf{7.01} & \textbf{7.83} & \textbf{4.07} & \textbf{20.77} & 9.13 \\
      & AC & \textbf{5.69} & 8.57 & 4.22 & 2.27 & 3.73 & 1.20 & 6.06 & 7.24 \\
      & FL & 3.79 & 3.97 & 3.02 & 3.60 & 2.90 & 1.90 & 7.97 & \textbf{10.67} \\
    \midrule
      & $\mu$ & 6.63 & \textcolor{blue}{7.75} & \textcolor{blue}{5.22} & 6.36 & 4.56 & \textcolor{blue}{5.08} & \textcolor{blue}{8.01} & \textcolor{blue}{4.42} \\
      & \textcolor{black}{$\sigma$} & \textcolor{black}{3.64} & \textcolor{black}{4.16} & \textcolor{black}{2.23} & \textcolor{black}{4.09} & \textcolor{black}{2.01} & \textcolor{black}{3.37} & \textcolor{black}{4.70} & \textcolor{black}{2.87} \\      
      & max & 13.53 & 19.91 & 10.22 & 14.84 & 9.8 & 12.9 & 20.77 & 10.67 \\
    \end{tabular}}
  \label{tab:ClinicalData}%
\end{table}%

\begin{figure}[t!]
    \centering
    \includegraphics[width=\linewidth]{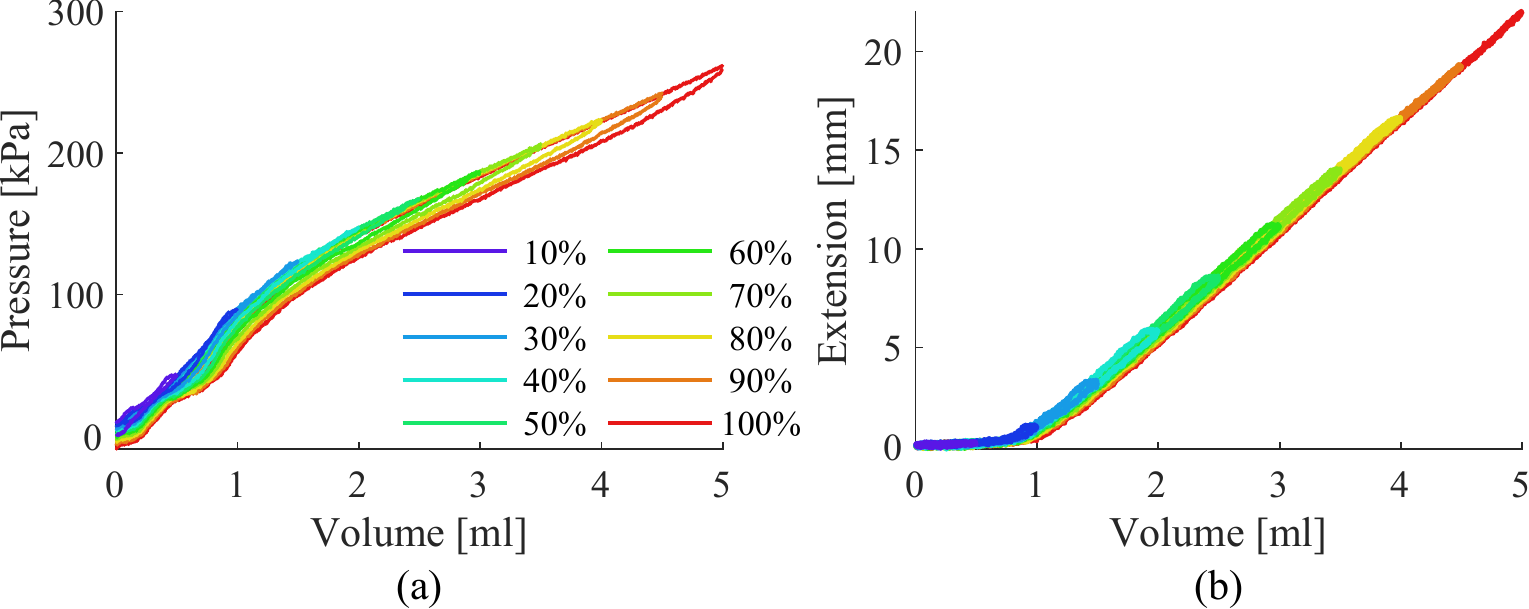}
    \caption{SFA pressure (a) and extension (b) under increasing working fluid volume \textcolor{black}{for different inflation levels}.}
    \label{fig:1DOF}
\end{figure}

\begin{figure*}
    \centering
    \includegraphics[width=\textwidth]{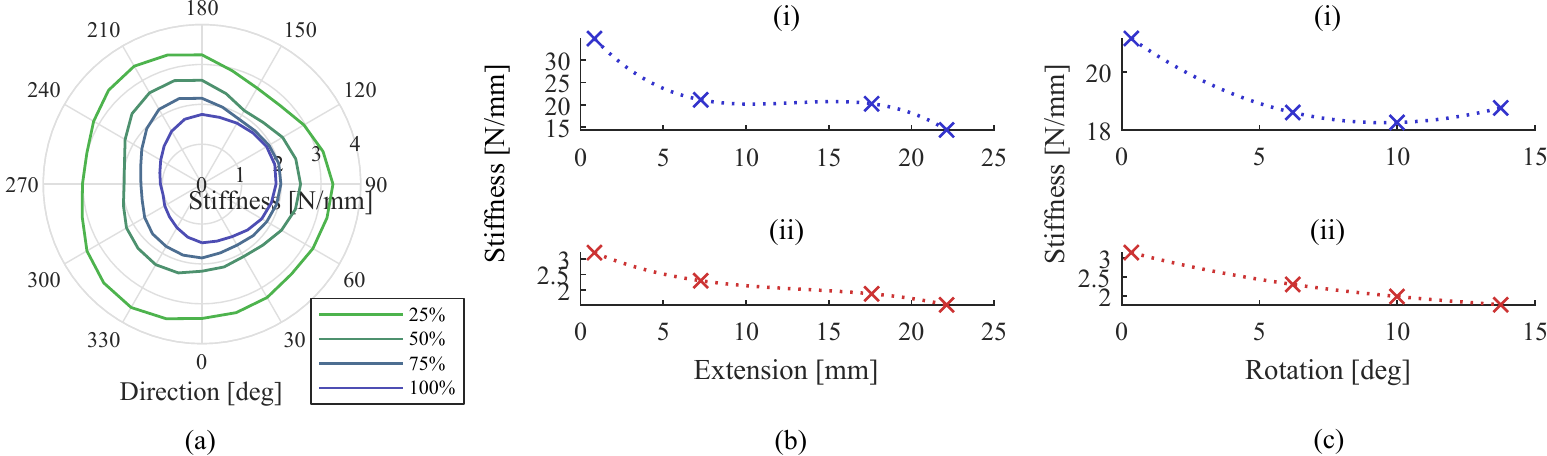}
    \caption{Change in transversal stiffness with the direction of the applied force for different extensions (a). Change in stiffness with extension for axial (i) and transversal stiffness (ii) (b). Change in stiffness with bending for axial (i) and transversal stiffness (ii) (c).}
    \label{fig:Stiffness}
\end{figure*}

\begin{figure}
    \centering
    \includegraphics[width=\linewidth]{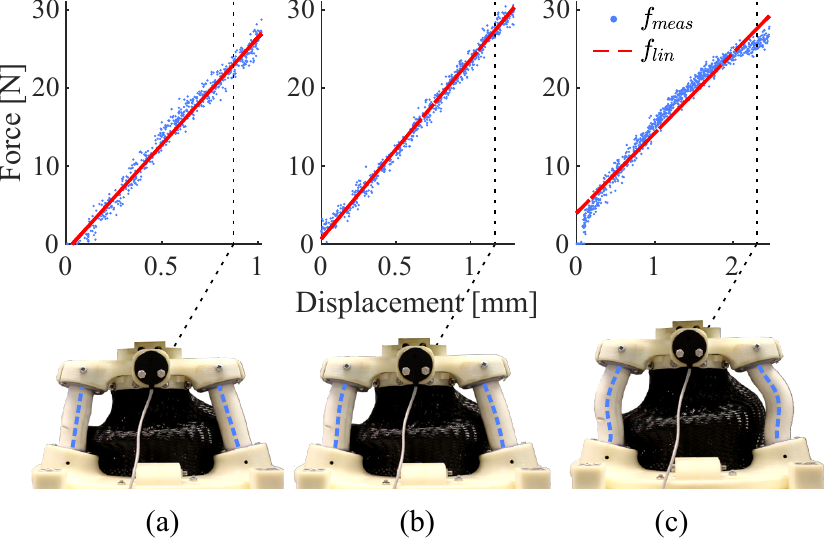}
    \caption{Measured compression force of the SEE $f_{meas}$ at 0\% (a), 50\% (b) and 100\% axial extension with it's corresponding linear interpolation $f_{lin}$. For each configuration the compressed SEE is depicted and the SFA centerlines are highlighted.}
    \label{fig:Buckling}
\end{figure}
\subsection{SFA characterization}
\label{SFA_characterization}
The results of the SFA characterization are shown in Fig. \ref{fig:1DOF}. The hydraulic pressure under SFA inflation and the resulting extension are shown in Fig. \ref{fig:1DOF}a) and \ref{fig:1DOF}b) respectively.
\textcolor{black}{Hysteresis is mainly observable in the fluid pressure. The mean deviation from the centerline between loading and unloading is 3.82$\pm$1.63kPa for the pressure and 0.14$\pm$0.05mm for the extension across the different inflation cycles. A maximum deviation due to hysteresis is observable in the pressure when inflated to 100\% with 9.28kPa and when inflated to 50\% at 0.44mm}.
The volume-extension curve of the SFA can be separated into two regions, a nonlinear (0ml to $\approx$1.25ml) and a linear region ($\approx$1.25ml to 5ml). In the linear region, the relationship can be can be approximated with a first order polynomial as $\Delta L(\Delta V)=6.61\text{mm}/\text{ml} - 5.52\text{mm}$. \textcolor{black}{The interpolation is used to determine the relationship between the SFA length change and the input volume change of the form $a = {\Delta V}/{\Delta L} = \pi \cdot 6.9^2 \text{mm}^2$}. As the proportion of the nonlinear region compared to the overall extension of the SFA is small, it is ignored for the following investigations. SFAs are therefore assumed to be pre-extended with a volume of 1.25ml.

\begin{figure*}[t!]
    \centering
    \includegraphics[width=\linewidth]{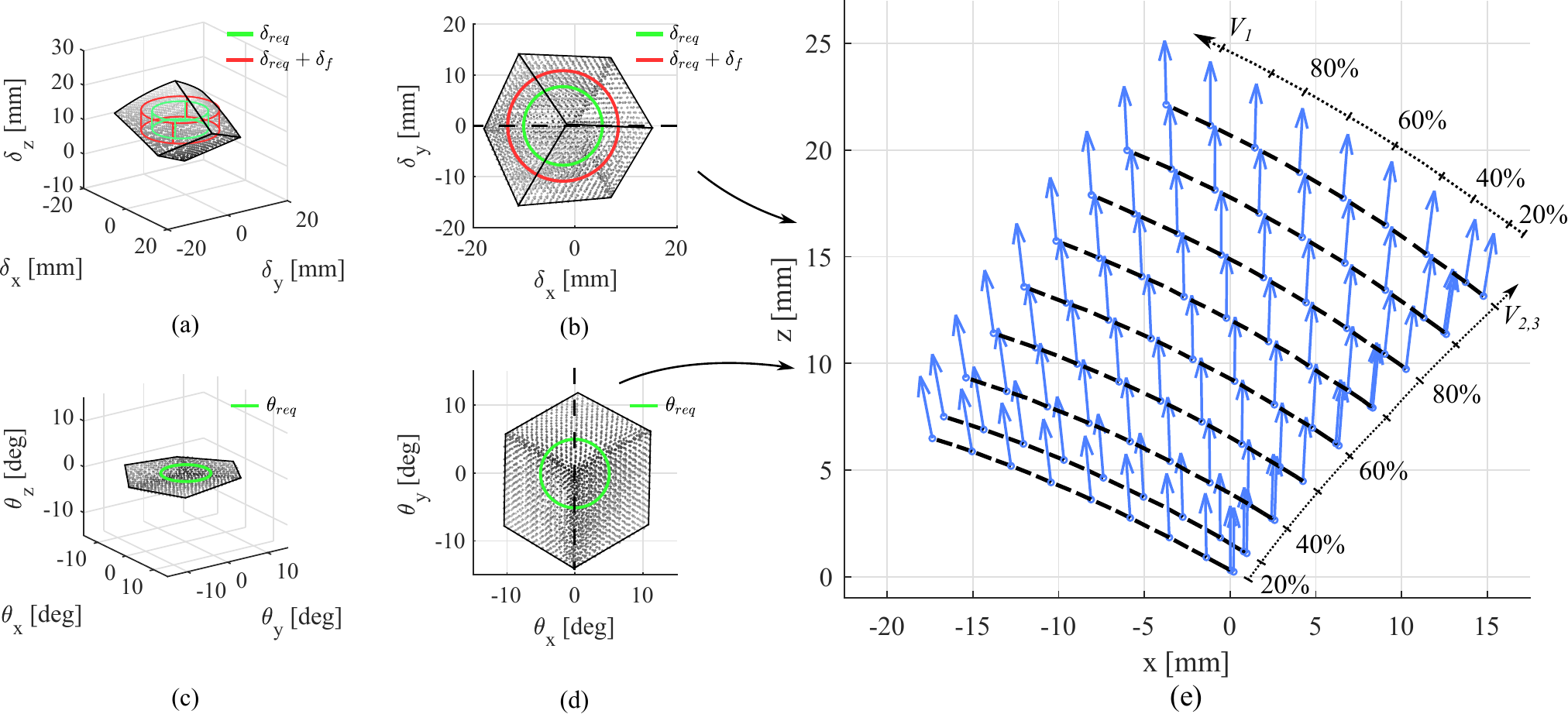}
    \caption{Workspace of the SEE in position (a-b) and orientation (c-d). \textcolor{black}{The required workspace $\boldsymbol{\delta}_{req}$ without and with consideration of the deflected tip pose $\boldsymbol{\delta}_f$ are indicated. A cross-section view along the dotted lines shows the coupling between position and orientation in the performed bending motions (e), in which the dashed lines indicate iso-volume lines for $V_2=V_3$.}}
    \label{fig:WSPos}
\end{figure*}

\subsection{Stiffness}
\label{Results_Stiffness}
The results of the twist stiffness characterization for each mesh configuration are shown in Table \ref{tab:Twist}, where $\mu$ and $\sigma$ are the mean and standard deviation of the twist stiffness $K_{tw}$ respectively. The application of a nylon mesh helps to significantly stiffen the torsional axis of the system by 184\%. A crimped mesh can further improve the torsional stiffness to 299\% of its original value. 

\begin{table}[h!]
\centering
\caption{Twist stiffness}
\label{tab:Twist}
\begin{tabular}{@{}cccc@{}}
\toprule
& None & Uncrimped & Crimped\\ \midrule
$\mu(K_{tw})$ [Nmm/$\degree$] & 45.94 & 84.68 & 137.37\\   
$\sigma(K_{tw})$ [Nmm/$\degree$] & 0.70 & 4.30 & 2.73 \\
\bottomrule
\end{tabular}
\end{table}
The results of the lateral stiffness characterization under inflation of the SEE are shown in Fig. \ref{fig:Stiffness}a) in polar coordinates. The radius indicates the magnitude of the stiffness in the given direction.

The axial and averaged lateral stiffness of the SEE under axial extension are presented in Fig. \ref{fig:Stiffness}b). The data are presented alongside their corresponding spline interpolations. Both decrease monotonically with the axial stiffness starting from a maximum of 34.83N/mm and reaching a minimum of 14.41N/mm at 100\% extension. The transversal stiffness decreases at a comparable rate from 3.21N/mm at 25\% down to 1.51N/mm at 100\% extension.
The stiffness variation under bending of the SEE is shown in Fig. \ref{fig:Stiffness}c) with the visualized trends interpolated by splines. Whilst the the transversal stiffness decreases monotonically from 3.15N/mm to 1.77N/mm, the axial stiffness decreases from 21.15N/mm at 0.3$\degree$ tilt to a minimum of 9.99N/mm at 10$\degree$ followed by an increase in stiffness to 18.75N/mm at 13.75$\degree$. The presented data is employed to determine the minimum stiffness of the system throughout the workspace to infer possiblly occurring tip pose deviations from external forces. It can be seen that the system reaches a minimum axial stiffness of 14.41N/mm and transversal stiffness of 1.51N/mm, both in a straight and fully extended configuration.

Despite high loads along the axial direction of the SEE no discontinuous buckling behaviour of the SFAs is observable. This is demonstrated in Fig. \ref{fig:Buckling}. The force-displacement relationships and their corresponding linear interpolations are shown for 0\%, 50\% and 100\% extension and depictions of the SEE at the corresponding maximum loads are presented. Whilst a slight increase in the nonlinearity between force and displacement is observable for 100\% extension (the corresponding mean absolute errors between data and linear interpolation are 0.84N, 0.62N and 1.16N for 0\%, 50\% and 100\% extension) no discontinuities are identifiable. The depictions of the deformed SEE show how the forced S-shape bending of the SFAs helps to prevent buckling. An increase in axial force only causes the curvature of the S-bend to increase.

\subsection{Workspace}
The workspace of the SEE in position and orientation is shown in Fig. \ref{fig:WSPos}. The figures show the tip pose acquired by the EM tracker for any given SFA configuration. The required workspace in position and orientation, $\boldsymbol{\delta}_{req}$ and $\boldsymbol{\theta}_{req}$, obtained in Section \ref{ClinicalData} from clinical data is projected into the center of the SEE workspace.
The deflected workspace $\boldsymbol{\delta}_f$ is calculated from the results obtained in Section \ref{Stiffness}. It can be seen that the SEE exhibits an minimum transversal stiffness of $1.51\text{N}/\text{mm}$ and a minimum axial stiffness of $14.41\text{N}/\text{mm}$ at 100\% extension. Taking into account the mean external load applied to the tip, a possible additional deflection of
\[
    \boldsymbol{\delta}_f =
    \begin{bmatrix}
    14.41& 0\\
    0& 1.51
    \end{bmatrix}^{-1}
    \begin{bmatrix}
    8.01\\
    4.42
    \end{bmatrix}
    =
    \begin{bmatrix}
    0.56\\
    2.95
    \end{bmatrix}
\]
Thus, the workspace the SEE is required to achieve extends correspondingly to
\begin{equation}
    \boldsymbol{\hat{\delta}}_{req} = \boldsymbol{\delta}_{req} + \boldsymbol{\delta}_f =
    \begin{bmatrix}
        5.78\\
        10.68
    \end{bmatrix}
\end{equation}
Whilst in some instances larger motions have to be achieved, the derived values represent a baseline motion range desirable from the SEE.

\textcolor{black}{To quantify whether the SEE is able to reach the desired workspace, the intersections between requirement and SEE workspace volumes are computed. It can be seen that for the unloaded requirements in translation and rotation, $\boldsymbol{\delta}_{req}$ and $\theta_{req}$, the SEE can accomplish 100\% of the workspace. For the workspace adapted to account for an external force $\boldsymbol{\hat{\delta}_{req}}$, the robot achieves 95.18\% of the required workspace.}

It is shown that a maximum combined lateral deflection of 19.01mm can be reached along the principal plane of $SFA_3$, which is about 4.5\% lower than the maximum transversal motion \textcolor{black}{observed in manual scanning}. The maximum extension of the SEE of 22.13mm is reached for a full inflation of all SFAs \textcolor{black}{and exceeds the demanded axial translation of 10.22mm as well as the transversal translation of 19.91mm determined from the clinical data}. The maximum tilt of the SEE is reached along the principal plane of $SFA_1$ with 14.02$\degree$\textcolor{black}{, which is $\approx9\%$ greater than the maximum demanded tilt of 12.9$\degree$}. A maximum axial torsion of 1.03$\degree$ occurs. Compared to the tilt ranges in X and Y the twist is significantly lower and will therefore be ignored in the following investigations.
\textcolor{black}{The coupling  between translation and rotation, the bending, of the SEE upon actuator inflation is shown in Fig. \ref{fig:WSPos}e) for a cross-section of the workspace along the central x-z-plane in translation and the corresponding y-axis of the rotational workspace. It can be seen that with the amount of transversal translation, the rotation of the tip increases, whilst axial extension has no effect on the rotation.}

\begin{table}[h!]
\centering
\caption{Repeatability}
\label{tab:Repeatability}
\begin{tabular}{@{}cccc@{}}
\toprule
Pose & \textcolor{black}{$[V_1, V_2, V_3]$}& $||\delta_e|| [\text{mm}]$ & $||\theta_e|| [\degree]$ \\
\midrule
\textcolor{black}{$C_1$}&$[0\%, 0\%, 0\%]$& $0.07 \pm  0.05$& $0.03  \pm  0.02$\\   
\textcolor{black}{$C_2$}&$[75\%, 50\%, 75\%]$& $0.10 \pm 0.05$ & $0.05 \pm    0.02$\\   
\textcolor{black}{$C_3$}&$[25\%, 0\%, 100\%]$& $0.07  \pm  0.03$ & $0.06  \pm  0.03$\\   
\textcolor{black}{$C_4$}&$[50\%, 25\%, 0\%]$& $0.08  \pm  0.06 $& $0.04  \pm  0.02$\\   
\textcolor{black}{$C_5$}&$[70\%, 80\%, 25\%]$& $0.09  \pm  0.04$ & $0.06 \pm  0.03$\\   
\textcolor{black}{$C_6$}&$[0\%, 20\%, 70\%]$&$ 0.11  \pm  0.05$ & $0.07  \pm  0.03$\\   
\midrule
\multicolumn{2}{c}{$\mu$ }  & 0.09 & 0.05\\
\bottomrule
\end{tabular}
\end{table}

The results of the positioning repeatability evaluation are presented in Table \ref{tab:Repeatability}. The table indicates the mean Euclidean errors in position and orientation with their respective standard deviations from the given pose for the 50 repetitions \textcolor{black}{with respect to the mean pose for the given configuration, $\mu(\boldsymbol{x}(C_j))$}. \textcolor{black}{For a configuration $C_j$, for instance, the Euclidean error $||\delta_e||$ is computed as
\begin{equation}
    ||\delta_e|| = \sum_{i=1}^{n=50}\frac{||\boldsymbol{x}_i-\mu(\boldsymbol{x}(C_j))||}{n}
\end{equation}
The pose $\boldsymbol{x}_i$ for a given configuration $C_j$ is obtained by averaging the measured static tip pose over a period 4 seconds. The orientation error $||\theta_e||$ and both corresponding standard deviations are calculated in the same manner.
Whilst it can be seen that the measured accuracy of the SEE is with $\approx 0.1$mm in position and $0.05\degree$ orientation slightly below the rated accuracy of the EM tracking system (0.48mm and 0.30$\degree$ RMS \cite{NDI2013}), it can be seen that averaging the pose data over 4 seconds reduces noise-related variance in the data. The samples are normally distributed across the workspace and thus the time-averaged mean is assumed to represent the tip pose sufficiently.}

\begin{figure}[t]
    \centering
    \includegraphics[width = \linewidth]{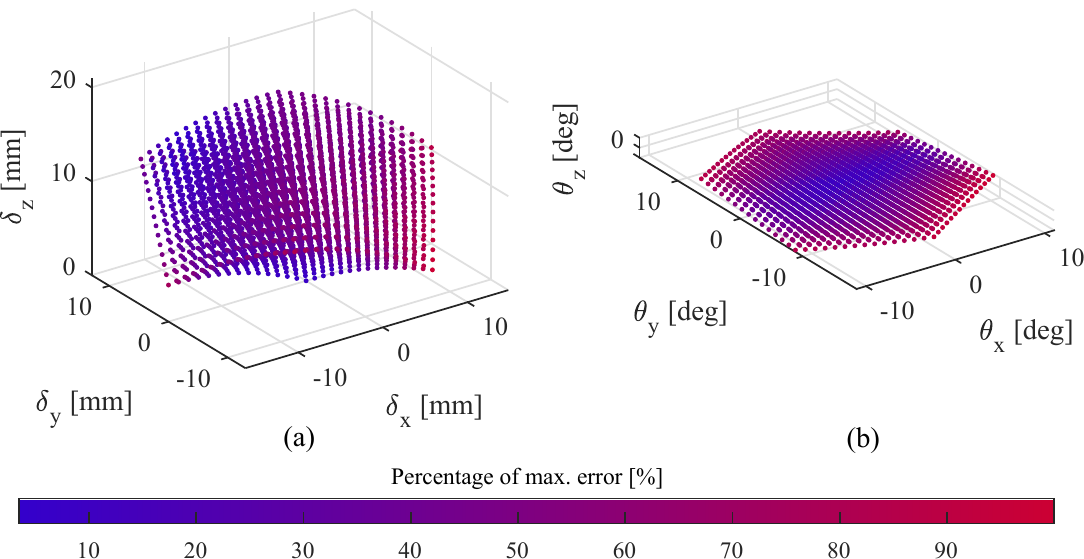}
    \caption{Workspace generated with model in position (a) and orientation (b). The colour indicates the normalized Euclidean error in the given state with respect to the maximum deviation from the model\textcolor{black}{, which is 2.37mm in position and 2.46$\degree$ in orientation}.}
    \label{fig:ModelValidation}
\end{figure}

\subsection{Model validation}
\label{Model_validation}
The results of the model validation are shown in Fig. \ref{fig:ModelValidation} and summarized in Table \ref{tab:ModelValidation}\textcolor{black}{, where $\mu$ refers to the mean error, $\sigma$ to the standard deviation and $max$ to the maximum error}. The estimated workspace of the SEE generated with the kinetostatic model is shown in Fig. \ref{fig:ModelValidation}. The colour of each marker indicates the Euclidean distance between the calculated point and the corresponding measured pose normalized to the maximum error in position and orientation respectively, namely 2.37mm and 2.46$\degree$. \textcolor{black}{The Young's modulus of the SFA material $E$ and its area moment of intertia $I$ have been manually tuned to minimize the Euclidean error in position and orientation. The obtained values are shown in Table \ref{tab:ModelParameters}.}

Overall, the model validation shows with a mean Euclidean error of $1.18\pm0.29$mm in position and $0.92\pm0.47\degree$ in orientation good results in predicting the tip pose under SFA extension.

\begin{table}[h!]
\centering
\caption{Model validation}
\label{tab:ModelValidation}
\begin{tabular}{@{}ccccccc@{}}
\toprule
 & \multicolumn{3}{c}{Displacement [mm]} & \multicolumn{3}{c}{Tilt [$\degree$]} \\ \cline{2-7}
 & $\mu$ & $\sigma $ & max & $\mu$ & $\sigma $ & max \\ \midrule
$e_x$ & -0.81 & 0.20 & 1.25 & 0.04 & 0.44 & 1.34 \\   
$e_y$ & -0.55 & 0.47 & 1.60 & 0.05 & 0.86 & 2.26 \\   
$e_z$ & -0.05 & 0.50 & 1.80 & -0.13 & 0.36 & 1.36 \\   
$||e||$ & 1.18 & 0.29 & 2.37 & 0.92 & 0.47 & 2.47 \\ 
\bottomrule
\end{tabular}
\end{table}

\begin{figure}[t]
    \centering
    \includegraphics[width=\linewidth]{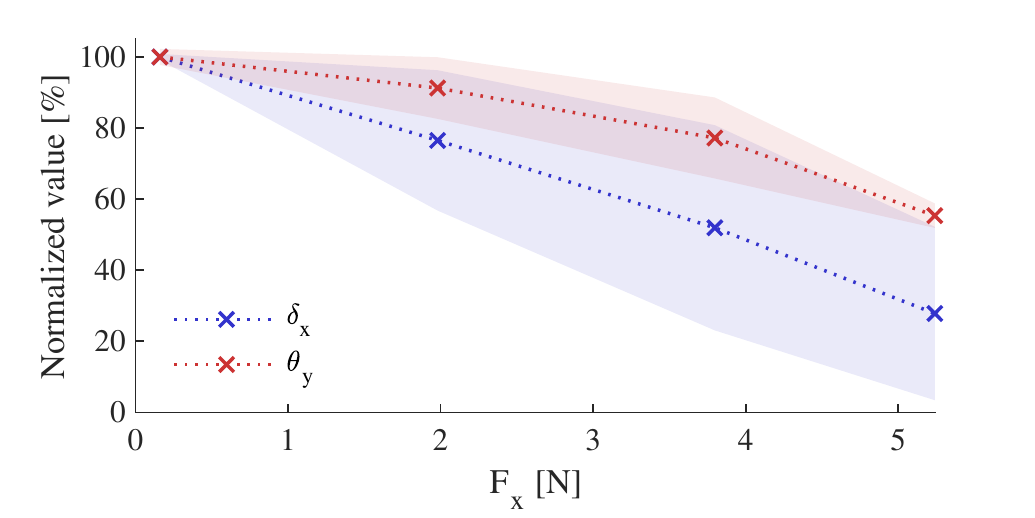}
    \caption{Effect of axial loading on transversal motion}
    \label{fig:Contact}
\end{figure}

\begin{figure}[t]
    \centering
    \includegraphics[width=\linewidth]{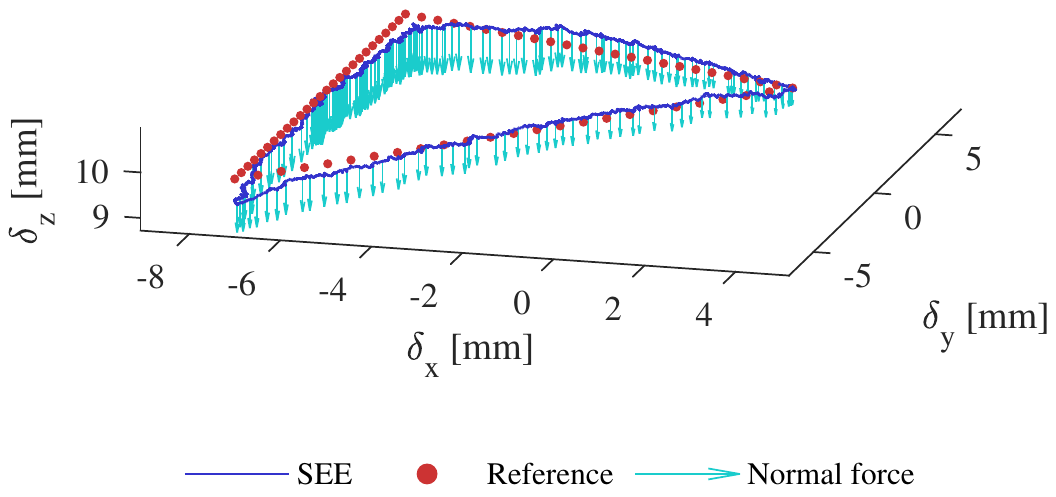}
    \caption{Example of tracked trajectory under external loading. The normal force is represented with scale of 0.2mm/N}
    \label{fig:PositionControl}
\end{figure}
\subsection{Contact experiment}
The motion constraint induced by an indentation contact is investigated. Fig. \ref{fig:Contact} shows the constraint of the mean X-displacement and Y-tilt for a given motion over 10 repetitions normalized to \textcolor{black}{their respective} maximum value\textcolor{black}{s of 12.92mm in position and 8.67$\degree$ in orientation when no external force is present, as well as their corresponding linear interpolations}. The transversal force applied by the SEE is measured with the force torque sensor. For both, the displacement and the tilt, the magnitude declines monotonically. Whereas the displacement reaches a minimum at 27.84\%, the tilt remains less affected by the lateral force with a minimum of 55.35\%. Linearizing the trends yield a decrease of 14.09$\%/\text{N}$ for the displacement and only 8.56$\%/\text{N}$ for the tilt. 

\begin{table}[t!]
\centering
\caption{Position control results}
\label{tab:PositionControl}
\begin{tabular}{@{}ccccccc@{}}
\toprule
 & \multicolumn{3}{c}{Flat - Unloaded}& \multicolumn{3}{c}{Tilted - Unloaded} \\ \cline{2-7}
 & $\mu$ & $\sigma$ & max & $\mu$ & $\sigma$ & max\\ \midrule
$e_x$ [mm] & 0.20 & 0.19 & 0.98 & 0.19 & 0.19 & 0.82\\   
$e_y$ [mm] & 0.25 & 0.20 & 1.03 & 0.27 & 0.20 & 0.88\\   
$e_z$ [mm] & 0.11 & 0.10 & 0.52 & 0.13 & 0.07 & 0.35\\   
$||e||$ [mm]  & 0.34 & 0.29 & 1.51 & 0.36 & 0.28 & 1.25\\ \midrule
& \multicolumn{3}{c}{Flat - Loaded}& \multicolumn{3}{c}{Tilted - Loaded}\\ \cline{2-7}
 & $\mu$ & $\sigma$ & max & $\mu$ & $\sigma$ & max\\ \midrule
$e_x$ [mm] & 0.24 & 0.23 & 1.10 & 0.27 & 0.28 & 1.50 \\   
$e_y$ [mm] & 0.33 & 0.22 & 1.08 & 0.32 & 0.25 & 1.07 \\   
$e_z$ [mm] & 0.13 & 0.10 & 0.56 & 0.17 & 0.10 & 0.65 \\   
$||e||$ [mm]  & 0.42 & 0.33 & 1.64 & 0.45 & 0.39 & 1.95 \\ \bottomrule
\end{tabular}
\end{table}

\subsection{Position control}
An example of a tracked trajectory with external loading is shown in Fig. \ref{fig:PositionControl}. The position controller tracks the desired position accurately with marginally larger tracking error around the corners of the triangular path. The quantitative results of the controller evaluation for the three executions are presented in Table \ref{tab:PositionControl} for both the unloaded and loaded trajectories, \textcolor{black}{where, as in Section \ref{Model_validation}, $\mu$ refers to the mean error, $\sigma$ to the standard deviation and $max$ to the maximum error in the respective direction}. The results indicate a higher mean error for the z-direction regardless of the configuration, which is also observable in the visualization above.

\begin{figure}[t]
    \centering
    \includegraphics[width=\linewidth]{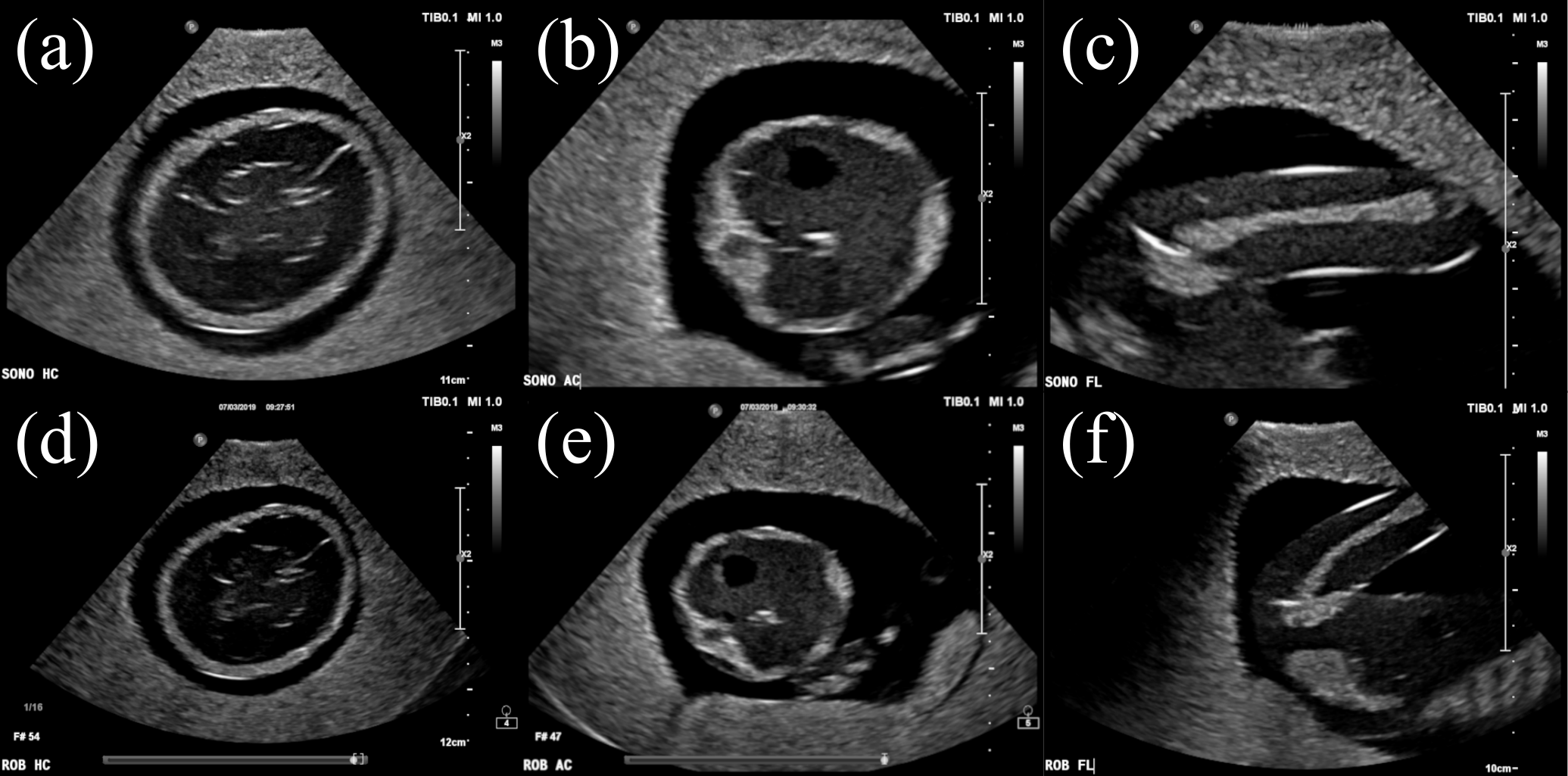}
    \caption{Ultrasound images acquired by sonographer (a-c) and SEE (d-f) for HC (a,d), AC (b,e) and FL (c,f) measurements }
    \label{fig:USImages}
\end{figure}

\subsection{Teleoperation and image-acquisition}
\label{ImageAcquisition}
The images obtained through manual ultrasound probe placement and steering with the SEE are presented in Fig. \ref{fig:USImages}. Anatomical structures of the foetus phantom are clearly visible throughout all images with minor shadowing on the left side of the FL standard view-plane, outside of the region of interest. In both cases, the regions of interest are centered in the image. Moreover, the contrast in the robot-acquired images is similar to the one in the manually-obtained images.

\section{Discussion}

In this work we developed a soft robotic ultrasound imaging system  \textcolor{black}{to offload sonographers in day-to-day scanning routines. The system addresses the issue of providing a stable contact between the ultrasound probe and patient, which could help improve sonographers’ ergonomics particularly with respect to work-related musculoskeletal disorders which arise from stresses induced by repeated manual probe handling.} The robot allows for tele-operated scanning and provides a platform for advanced imaging approaches. It is designed in form of an end-effector which is manually positioned in the area of interest and actively steered towards the desired view-plane. Due to its inherent compliance, the SEE is able to maintain contact while exhibiting sufficient axial stiffness to ensure mechanical coupling for the ultrasound image acquisition \textcolor{black}{which is verified by acquiring standard views on a foetal ultrasound phantom}.

The system shows with its high axial and low lateral stiffness good applicability to foetal ultrasound scanning. Despite the quick decline of stiffness with axial extension, the SEE is with $14.41\text{N}/\text{mm}$ axial stiffness at full extension still capable to apply sufficiently high forces to the patient without significant deformation, \textcolor{black}{which is approximately 1.44mm at maximum axial load of 20.77N}. The lower lateral stiffness allows for the system to be adaptable to the contact surface and to be moved away in case of discomfort in the patient \textcolor{black}{whilst being sufficiently high to counteract transversal loads occurring during the intervention. It can be seen that for the fully extended SEE the transversal displacement at a maximum occurring load of 10.67N reaches 7.1mm.}

\textcolor{black}{The compliance of the system allows for deformation upon external motion when clamped onto a patient. Thus, the resulting contact force is significantly lower compared to a rigid system. It furthermore exhibits a low mass which could be beneficial in the dynamic safety of the system \cite{Haddadin2008}.}

\textcolor{black}{If the stiffness in the axial direction of the probe needs to be adjusted or the tip force controlled, the system can be equipped either with a force sensor at the base to estimate tip forces or serve as a sensor itself \cite{Lindenroth2017IntrinsicActuation}. While in the first case the tip pose change during the operation needs to be accounted for to accurately determine the external force, either by an accurate model or pose feedback, the second case can make use of the deformable structure of the robot paired with the kinematic constraints induced by the actuation channels to infer the external force.}

We have shown that the integration of a braided nylon mesh, which has previously only been used to avoid ballooning in SFAs, can significantly improve the twist stiffness of the SEE to up to three times in comparison to the mesh-free system. \textcolor{black}{The use braided meshes is a highly versatile design approach and shows the potential to become a de facto standard in reinforcing not only soft robotics system but also continuum robots against unwanted external twists induced by contact wrenches, thus enabling such robots for a wider range of applications.}

\textcolor{black}{The workspace achieved by the SEE covers without external loading the \textcolor{black}{average} translation and rotation motion ranges required to achieve a desired view, as shown from clinical data. Loading the probe with the contact forces measured in clinical scans and assuming the lowest possible stiffness of the system reduces the achieved workspace to about 95.18\% of the \textcolor{black}{mean required range}. Whilst\textcolor{black}{, for example,} the maximum translation of the SEE is at 19.01mm significantly higher than the required deflected motion of 10.68mm, the non-homogeneous shape of the SEE workspace dictates the limitations in covering the required translation range.} This limitation could be addressed by adding a linear stage to the base of the SEE to allow for axial translation without sacrificing the softness of the system. \textcolor{black}{Moreover, an axial rotation stage could be added to allow for more complex probe motions.}

\textcolor{black}{A high variability in the monitored ultrasound probe motion ranges can be observed across the obtained views and subjects. Whilst, on average, relatively small maximum deflections are observed, in some instances significantly larger motions occur. This is indicated by the high standard deviations in the motion ranges of the respective axes. Further research needs to be conducted into the exact metrics of the ultrasound probe motions and whether the designed system can satisfy those metrics. Additional considerations such as the coupling between different motion axes then need to be accounted for. Another factor in the feasibility of a desired view is the accuracy of the manual placement of the passive positioning arm. If the accuracy is low and the view is out of reach of the end-effector, the passive arm could either be repositioned manually or additional DOFs could be added to the system. More accurate methods should be employed in evaluating the manual probe motions. The use of a percentile is difficult for the given data due to the high variability in the times required to obtain desired views, as seen in the presented time series for the motions of subject 5 in Fig. 10 for example. Thus, a larger scale and more streamlined data acquisition needs to be conducted.}

\textcolor{black}{We showed that the combination of SFAs and hydraulic actuation exhibits good properties for the SEE to be driven in an open-loop configuration. The relationship between SFA length and input volume is highly linear and only shows 0.14$\pm$0.05mm deviation due to hysteresis, thus allowing for an accurate prediction of the kinematic constraints imposed on the SEE. This compliments the derived kinetostatic model, which is able to accurately predict the SEE tip motion with an accuracy of 1.18mm in position and 0.92$\degree$ in orientation as a function of the induced working fluid volume. The model deviates more along the boundaries of the workspace, which could be caused by the larger deflection of the SFAs and resultant nonlinearieties caused by the bending of the actuators. This could be addressed by extending the model to a nonlinear approach, as we have for example demonstrated in \cite{Lindenroth2017IntrinsicActuation} for a soft continuum robot.}

\textcolor{black}{The repeatability lies with 0.1mm in position and 0.05$\degree$ in orientation slightly below the rated accuracy of the measurement system. As the obtained measurements are  expressed relative to a mean pose, averaged over time and normally distributed, it is assumed that these values still represent the true pose well. The high repeatability and should allow for accurate positioning of the SEE in view-plane finding applications.}

\textcolor{black}{The system maintains stability and controllability well when in contact with a tissue-like soft silicone rubber patch. We showed that the implemented closed-loop position controller is able to track target trajectories accurately with a mean position error of 0.35mm with only marginally increased errors in the tracking accuracy of 0.44mm when a contact force applied. In scenarios where EM tracking is not available, the ultrasound image could be used to provide pose feedback. This could then employed as a substitute for the position feedback in the closed-loop controller.}

The coupling between position and orientation is an obvious limitation in the usability of the design. It can be seen, however, that the mechanical properties of the surface contact greatly affect the coupling behaviour. We have shown that an indenting contact reduces the lateral motion of the ultrasound probe significantly more than the tilt. It can easily be seen that a very stiff coupling in combination with the minimal contact friction caused by the application of ultrasound gel greatly reduces the tilt capabilities of the system while allowing for lateral sliding. It can therefore be assumed that in practice the coupling can be reduced by varying the axial pressure applied to the patient. This is supported by the findings of the tele-operated image acquisition in Section \ref{ImageAcquisition} and will be investigated further in future research.

\section{Conclusion}
The SEE design proposed in this work \textcolor{black}{shows a novel approach to applying soft robotics technologies in medical ultrasound imaging.} We have shown that under certain conditions the SEE satisfies the requirements imposed by the clinical application. The derived kinetostatic model mimics adequately the behaviour of the physical robot and the integrated system is capable of tracking target trajectories accurately and obtaining high-quality ultrasound images of a prenatal ultrasound phantom. In our future work, we will make use of the hydraulic actuation to integrate a force-controlled system through intrinsic force sensing, as shown in our previous work \cite{Lindenroth2017IntrinsicActuation}.

\begin{figure}[h!]
    \centering
    \includegraphics[width=\linewidth]{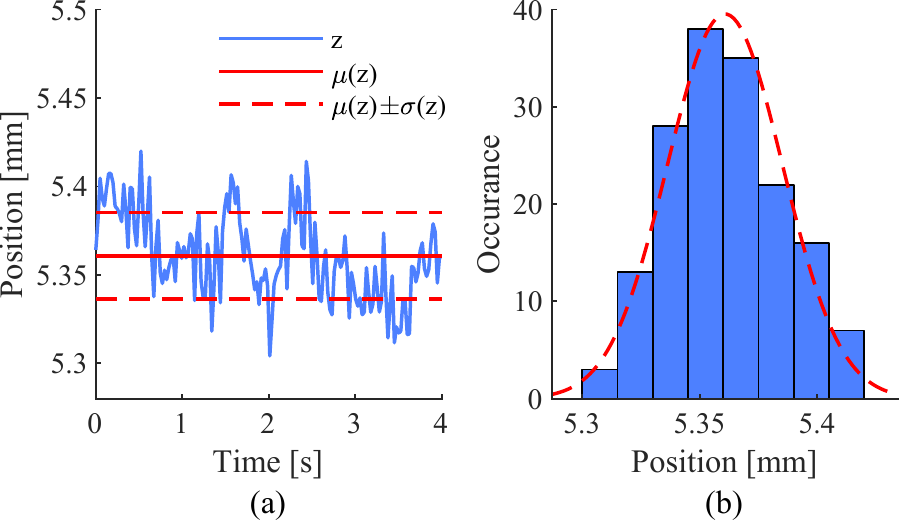}
    \caption{Sampling of EM tracker data for defined pose over 4sec (a) and corresponding distribution of measurements (b).}
    \label{fig:Rep1}
\end{figure}

\appendix[Error behaviour in repeatability validation]
For each achieved configuration in the repeatability validation experiment, the pose data is averaged over a period of 4 seconds. The resulting data for the displacement in Z-direction upon reaching configuration $C_4$ is presented in Fig. \ref{fig:Rep1}a). The corresponding distribution of measurements is shown in Fig. \ref{fig:Rep1}b). It can be seen that the readings follow a normal distribution around a mean of 5.36mm with a standard deviation of 0.03mm. A $\chi^2$ goodness-of-fit is performed to determine the suitability of describing the individual readings for a given pose as a normal distribution. Across all configurations, the mean $p$ value associated with the fit is 0.34$\pm$0.28. It is therefore concluded that this hypothesis holds across the workspace and thus, the time-averaged pose is a suitable indicator for the true pose of the SEE.

\section*{Safety considerations}
The use of a soft robotic system can help to greatly reduce the contact contact forces upon undesired patient or motions of the robot itself.

The build-up of contact force with a clamping contact between robot and the patient constrained by the patient bed can lead to discomfort and potentially injury \cite{Haddadin2008a}. To determine an approximate occurring force for a patient motion of 1cm against a rigid robot, we can calculate the following.
The Young’s modulus for visceral contents can be approximated by $E_{vis} = 8.42$kPa \cite{huang1994finite}. Assuming a circular contact of 10mm radius ($r$) with a tissue thickness ($d$) of 10mm, the stiffness of the visceral contents can be determined as
\[
K_{vis}={E \pi r^2}/d=39.37 \text{N}/\text{mm}
\]
If the patient moves against a stationary rigid robot over the distance $\Delta x=10$mm, the contact force experienced by patient and robot is
\[
f_{rigid}= K_{vis}\cdot \Delta x=39.37\text{N}/\text{mm}\cdot10\text{mm}=393.7\text{N}
\]
For the soft robot, the system stiffness is combined in form of two serially-connected springs. In case of the lowest transversal stiffness of the soft robot ($K_{SEE}=1.51 $N/mm), one can compute for the combined stiffness
\[
K_{comb}=(1/K_{vis} +1/K_{SEE} )^{-1}=1.45 \text{N}/\text{mm} 
\]
The resulting force build-up upon contact is then only 21.69N.
Considering the reduction in contact force when exposed to an involuntary patient or clinician motion, it can be assumed that the use of soft robots instead of rigid ones could greatly reduce contact forces when a patient is exposed to a clamping contact.

\section*{Acknowledgment}
This work was supported by the Wellcome Trust IEH Award [102431], the iFIND project, and by the UK Engineering and Physical Sciences Research Council (EPSRC) grant EP/R013977/1.

\ifCLASSOPTIONcaptionsoff
  \newpage
\fi

\bibliographystyle{unsrt2authabbrvpp}

\bibliography{references.bib}
\end{document}